%% file: main.tex
\documentclass{article}

\usepackage{microtype}
\usepackage{graphicx}
\usepackage{subfigure}
\usepackage{booktabs}

\usepackage{amsmath,amssymb,amsfonts,nicefrac}
\usepackage{amstext}
\usepackage{pgfplots}
\pgfplotsset{compat = 1.3}
\usepackage{multirow}
\usepackage{makecell}
\usepackage{booktabs}
\usepackage[ruled,vlined]{algorithm2e}
\usetikzlibrary{pgfplots.groupplots}

\usepackage{eucal}
\usepackage{ mathrsfs }

\usepackage{arxiv}

\def\x{{\mathbf x}}
\def\y{{\mathbf y}}

\def\L{{\cal L}}
\def\ngrams{\textit{n-grams}}

\def\tco{\texttt{train-clean-100}}
\def\tct{\texttt{train-clean-360}}
\def\tof{\texttt{train-other-500}}

\def\devother{\texttt{dev-other}}

\def\testother{\texttt{test-other}}
\def\llten{\texttt{LibriLight-10}}

\newcommand{\librivox}{\textsc{LibriVox}}
\newcommand{\librispeech}{\textsc{LibriSpeech}}
\newcommand{\librilight}{\textsc{LibriLight}}

\title{Iterative Pseudo-Labeling for Speech Recognition}

\author{
Qiantong Xu, Tatiana Likhomanenko, Jacob Kahn, Awni Hannun, Gabriel Synnaeve, Ronan Collobert \\
Facebook AI Research, Menlo Park \& New York, USA \\
\texttt{\{qiantong,antares,jacobkahn,awni,gab,locronan\}@fb.com} \\
}
\begin{document}

\maketitle
\begin{abstract}

Pseudo-labeling has recently shown promise in end-to-end automatic speech recognition (ASR). We study Iterative Pseudo-Labeling (IPL), a semi-supervised algorithm which efficiently performs multiple iterations of pseudo-labeling on unlabeled data as the acoustic model evolves. In particular, IPL fine tunes an existing model at each iteration using both labeled data and a subset of unlabeled data. We study the main components of IPL: decoding with a language model and data augmentation. We then demonstrate the effectiveness of IPL by achieving state-of-the-art word-error rate on the \librispeech{} test sets in both standard and low-resource setting. We also study the effect of language models trained on different corpora to show IPL can effectively utilize additional text. Finally, we release a new large in-domain text corpus which does not overlap with the \librispeech{} training transcriptions to foster research in low-resource, semi-supervised ASR.

\end{abstract}

\section{Introduction}
\label{sec:intro}
Recent advances in end-to-end speech recognition are largely due to acoustic model (AM) architecture improvements. Some of the most promising are from the Transformer family~\cite{synnaeve2019end,mohamed2019transformers,karita2019comparative,han2019stateoftheart,wang2019transformerbased}, which give state-of-the-art results on many ASR benchmarks and close the gap between end-to-end and hybrid systems. Given the performance gain from new architectures, research has shifted focus towards leveraging self- and semi-supervised techniques to better utilize unlabeled data. For example, pseudo-labeling successfully boosts the performance on \librispeech{}~\cite{panayotov2015librispeech} baselines by a large margin~\cite{synnaeve2019end}. Many algorithms exist which incorporate unlabelled data to improve ASR in the low-resource setting, including representation learning~\cite{librilight, baevski2019effectiveness, ling2020deep}, pseudo-labeling~\cite{kahn2019self}, local prior matching~\cite{hsu2020semi}, pseudo-label augmentation~\cite{chen2020semi}, adversarial training~\cite{liu2019adversarial} and back translation~\cite{baskar2019self}. While many of these methods outperform a supervised baseline with limited resources, a large gap to fully-supervised training remains. Furthermore, not all approaches scale easily to large amounts of data, such as that recently used in the \librilight{} benchmark~\cite{librilight}. 

\begin{figure}[t]
\centering
    \input{figures/motivation.tikz}
\caption{WER on \devother{} for two different strategies of using the unlabeled data. In fine tuning, the model is first trained on \tco{} only for 160 epochs and then used to generate pseudo-labels on \tct{}. Both \tco{} and \tct{} are then used to fine tune the same model. 
In "from scratch training", same pseudo-labels on \tct{} together with the true labels on \tco{} are used to train a new model with the same architecture from scratch.}
\label{fig:motivation}
\end{figure}
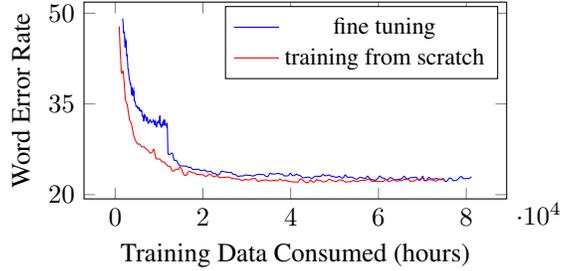

In this work, we study iterative pseudo-labeling (IPL), a straightforward method that can easily scale to large unlabeled datasets and further boost the performance in both standard- and low-resource settings. IPL is motivated by the simplicity and effectiveness of pseudo-labeling (PL)~\cite{kahn2019self,synnaeve2019end}. A simple extension to~\cite{synnaeve2019end} involves conducting more iterations of PL as the model trains so as to continuously refine and improve the quality of generated pseudo-labels.
That said, training a model from scratch after each round of pseudo-labeling and relabeling a large collection of unlabeled data is expensive.
IPL mitigates these challenges by 1) labeling only a subset of the unlabeled data in each iteration, and 2) fine tuning the existing model on this subset, rather than training from scratch. An intuitive motivation for this is shown in Figure~\ref{fig:motivation}, where the same acoustic model is trained to convergence with a fixed learning rate; both settings reach a similar word-error rate (WER). Training from scratch with PL is also shown to be roughly equivalent to iterating in a machine translation~\cite{he2019revisiting} and a image classification~\cite{xie2019self} setting.

\section{Related Work}
Semi-supervision in ASR is well-studied~\cite{wessel2004unsupervised,lamel2002lightly,ma2008unsupervised,yu2010unsupervised,manohar2015semi,liao2013large}; our work builds primarily on recent work with end-to-end systems, especially PL~\cite{kahn2019self, synnaeve2019end}. In~\cite{kahn2019self}, PL is shown to be effective with only 100h of labeled audio. The model uses a sequence-to-sequence loss and requires additional pseudo-label filtering to achieve the best results. To mitigate the instability found in sequence-to-sequence decoding, as in~\cite{synnaeve2019end} all pseudo-labels are generated with models trained with Connectionist Temporal Classification (CTC) loss~\cite{graves2006connectionist}. Our work extends~\cite{synnaeve2019end} by conducting more rounds of PL and fine tuning the existing model and demonstrates the effectiveness of the IPL approach in settings with both 960h and 100h of labeled audio. Still other work~\cite{baevski2019effectiveness} learns discrete audio feature representations directly from the waveform and works quite well with limited data, even in settings with under 100h of labeled audio. In this setting, learned acoustic features are presumably more competitive than an acoustic model trained end-to-end with MFCC or log-mel filterbank features. Other work including~\cite{hsu2020semi}, those with CPC baselines~\cite{librilight}, and those using adversarial training~\cite{liu2019adversarial} and back-translation-style techniques~\cite{baskar2019self} also provide promising end-to-end semi-supervised approaches, but results are not comparable as newer end-to-end approaches outperform these works' baselines.

\section{Method}
In this section, we first introduce the iterative pseudo-labeling algorithm (IPL). 
We then give theoretical justifications why IPL facilitates effective training.
Finally, we perform analysis and experiments on a small-scale labeled dataset.

\subsection{Iterative Pseudo-Labeling}
\label{sec:methods}

\begin{algorithm}[h!]
\SetAlgoLined
\KwData{Labeled data $L = \{x_i, y_i\}^l_{i=1}$, Unlabeled data $U = \{x'_j\}^u_{j=1}$}
\KwResult{Acoustic model $p_\theta$}
Initialize $p_\theta$ by training on only labeled data $L$\;
  \Repeat{convergence or maximum iterations are reached}{
    1. Draw a subset of unpaired data $\tilde U \in U$\;
    2. Apply $p_\theta$ and decoding with LM to the subset $\tilde U$ to generate $\hat U = \{(x, \hat y) | x \in \tilde U\}$\;
    3. Fine tune $p_\theta$ on $L \cup \hat U$ with data augmentation;
  }
 \caption{Iterative pseudo-labeling}
\label{algo:ipl}
\end{algorithm}

As listed in Algorithm \ref{algo:ipl}, IPL utilizes both labeled and unlabeled data as in the conventional semi-supervised learning. The model minimizes the following loss function:
\begin{equation}
    \mathscr{L} = \mathscr{L}_L + \lambda \mathscr{L}_U \label{eq:loss}
\end{equation}
where $\mathscr{L}_L$ and $\mathscr{L}_U$ denote the parts of the loss function on labeled and unlabeled data accordingly:
\begin{align}
    \mathscr{L}_L &= - \displaystyle \mathbb{E}_{\x,\y \sim p(\x,\y)} \log (p_{\theta}(\y|\x)) \label{eq:loss_p}\\
    \mathscr{L}_U &= - \displaystyle \mathbb{E}_{\x \sim p(\x)} \mathbb{E}_{\hat \y \sim p_{\theta}(\y|\x)} \log  (p_{\theta}(\hat \y |\x)). \label{eq:loss_u}
\end{align}
Note that in ASR, instead of sampling from $p_{\theta}(\y|\x)$, the transcriptions as well as the pseudo-labels are usually selected from the greedy path: 
\begin{equation}
\label{eq:greedyy}
\hat \y = \underset{\y}{\operatorname{argmax}} ~ p_{\theta}(\y|\x).
\end{equation}

\subsection{Avoidance of Local Minima}\label{sec:locmin} 
As discussed in~\cite{he2019revisiting}, one bane of loss (\ref{eq:loss}) optimization in fine tuning is that it tends to get stuck at a local minima after each round of training with existing pseudo-labels; the conditional log likelihood (\ref{eq:loss_u}) is already maximized when $p_{\theta}(\y |\x)$ matches the underlying data
distribution $p_{\theta^*}(\y |\x)$, so that $\nabla_\theta \mathscr{L} |_{\theta = \theta*} = 0$. The IPL algorithm has two distinct components: one with respect to the target ($\y$) and the other with respect to the data ($\x$), that we found to be effective to overcome this behaviour.

\subsubsection{External Language Model}
In modern ASR systems, in addition to the acoustic model $p_\theta$, a decoding procedure (typically either WFST-based~\cite{mohri2002weighted} or beam-search-based (BS)~\cite{pratap2018wav2letter,zeghidour2018fully} as well as lattice/beam rescoring) is always used to integrate an external language model (LM).
Thus, instead of using the greedy path (\ref{eq:greedyy}) as transcriptions, we consider:
\begin{equation}
\hat \y' = \underset{\y}{\operatorname{argmax}} ~ \log p_{\theta}(\y|\x) + \alpha \log p_{\text{LM}}(\y) + \beta |\y|, \label{eq:decoding}
\end{equation}
where $\alpha$ and $\beta$ are hyper-parameters \cite{synnaeve2019end} usually optimized on validation set. This differs $\hat \y'$ from $\hat \y$ by introducing extra LM knowledge into transcriptions so that the learned weights $\theta$ are no longer optimal given the new labels $\hat \y'$, and the model will keep training with $\nabla_\theta \mathscr{L} |_{\theta \neq \theta*} \neq 0$. This is also observed in machine translation~\cite{he2019revisiting}, where the gain from using greedy-path decoding is limited in self-training with PL.

\subsubsection{Data Augmentation}
With respect to data, when data augmentation is introduced, the log likelihood the model optimizes also changes. We can rewrite (\ref{eq:loss_u}) as
\begin{equation}
\mathscr{L}_U = - \displaystyle \mathbb{E}_{\x \sim p(\x), \x' \sim q(\x'|\x)} \mathbb{E}_{\hat \y \sim p_{\theta}(\y|\x)} \log (p_{\theta}(\hat \y|\x')),
\end{equation}
where $q(\cdot)$ is the data augmentation function, which is SpecAugment~\cite{park2019specaug} in our experiments. The model weights optimized before could be no longer optimal given the new augmented input; the model keeps updating with $\nabla_\theta \mathscr{L} |_{\theta \neq \theta*} \neq 0$.
\begin{figure}[t]
\centering
    \input{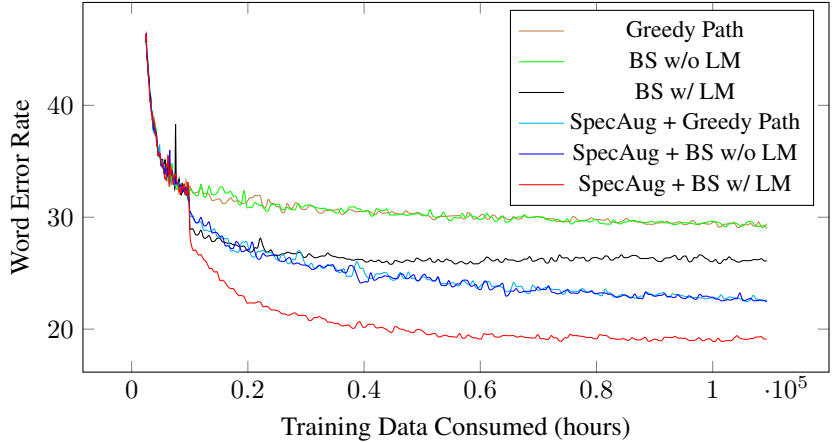}
\caption{WER on \devother{} with different IPL training strategies. Beam-search (BS) decoding is performed with an 4-gram LM.}
\label{fig:gain}
\end{figure}

\subsubsection{Empirical Study}
We use \tco{} and \tct{} in \librispeech{} as labeled and unlabeled training data and \devother{} as a validation set. The AMs detailed in Section \ref{sec:acoustic_models}, are first trained on labeled data for 100 epochs ($\approx 10^4$ hours) and then continue with IPL. Pseudo-labeling is performed every 10 epochs with all unlabeled data without down-sampling. The learning rate is fixed throughout training for a fair comparison. As shown in Figure~\ref{fig:gain}, if both data augmentation and LM decoding are used, there is a dramatic WER drop when unlabeled data is first in use, and the WER keeps decreasing as IPL progresses. If either data augmentation or LM decoding is removed, convergence degrades noticeably. 
Further, if both are removed, adding unlabeled data provides no benefit to IPL, which is in consistent with hitting local minima.
One should note that the model is not fully converged at epoch 100; the WER continues decreasing. The contribution of the beam search alone is also limited. 

\begin{figure}[t]
\centering
    \input{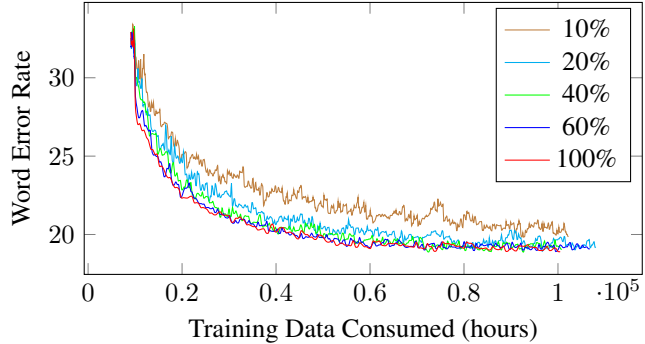}
\caption{WER on \devother{} with different amount of unlabeled data used in each iteration of IPL training (data augmentation and BS decoding with 4-gram LM are applied).}
\label{fig:px_estimation}
\end{figure}

\subsection{Dataset Distribution Approximation}
\label{sec:px_estimation}
Usually, unlabeled dataset $U$ is much larger than the labeled one~$L$; it is very time-consuming to label the entire~$U$ in each round of PL. If only insufficient data is selected in each round, however, the $p(\x)$ in (\ref{eq:loss_u}) will be poorly estimated, which will increase the chance of model overfitting. To balance the trade-off between the accuracy of $p(\x)$ approximation and the PL efficiency, with the same setup as in Section~\ref{sec:locmin}, we conduct an empirical ablation where the PL is performed only on a randomly sampled subset of $U$. Note that we treat samples from $L$ and $U$ equally following~\cite{synnaeve2019end}, so that the $\lambda$ in (\ref{eq:loss}) is implicitly set to $|U| / |L|$, the ratio between the number of samples in the two sets. 

As shown in Figure \ref{fig:px_estimation}, even though there is an up to 5-time gap in $\lambda$, using 20\% to 40\% of $U$ can reach the same WER as using 100\%. This motivates us to 1) pay less attention to $\lambda$ tuning and (2) down sample $U$ in PL, so as to perform more rounds of PL in total to better utilize large unlabeled set. On the other hand, using only 10\% from $U$ significantly hurts the convergence, which shows the importance of $p(\x)$ approximation and sets up a lower bound of down-sample rate.   

\section{Experiments}
\label{sec:experiments}

\subsection{Audio Data}
Audio data for our experiments comes from two sources: \librispeech{}, containing 960h of audio and paired transcriptions, and audio from \librivox{} (54K hours of audio) extracted following~\cite{librilight}. Three setups of labeled data are used: 1) full \librispeech{} (960h), 2) \tco{} subset (100h) from \librispeech{} and 3) the 10-hour training subset from \librispeech{} prepared in~\cite{librilight}, \llten{}. We use the standard development (for all hyper-parameters optimization) and test (for final evaluation only) sets from \librispeech{}. 

\subsection{Gutenberg Text Corpus}

As discussed in~\cite{kahn2019self}, it is important to remove components of the LM training corpus that overlap with unlabeled audio to ensure the LM has no information about ground truth transcriptions from the unlabeled audio. We study the contribution of the LM to IPL and conduct rigorous experiments when a subset of \librispeech{} is used as unlabeled audio. For LM training, we prepare a larger in-domain text corpus using books from Project Gutenberg \cite{gutenberg}. 
To prepare the corpus, we first start with a large subset of English books from Project Gutenberg (which includes some of the 14.5k books present in the \librispeech{} LM corpus~\cite{panayotov2015librispeech} with 0.8B words) and filter out all books present in \librivox{} audio data. We perform the same procedure as in~\cite{synnaeve2019end} along with a manual matching step to find exact or similar titles (after normalization) in \librivox{} ($\alpha$, $\beta$ are the same as in~\cite{synnaeve2019end}), filtering out the resulting books. Similarly, we remove from the corpus books present in the \librispeech{} validation and test sets. The resulting filtered corpus is normalized in the same way as in~\cite{synnaeve2019end} which mimics the normalization in the \librispeech{} corpus, but has additional mappings of some abbreviations and does not split text mid-sentence.
We denote this final corpus as $GB \setminus LV$ (2.16B words from 34k books).
Further, with the same procedure, we filter out books containing \librispeech{} training transcriptions (960h) and form a new corpus denoted as $GB \setminus LV \setminus LS$ (2.11B words from 33.4k books). 

\subsection{Models}

\subsubsection{Acoustic Model}
We use the best-performing Transformer architecture on \librispeech{} and \librivox{} with 322M parameters from~\cite{synnaeve2019end} in our experiments. In particular, there is a convolutional front-end containing 6 layers of 1-D convolutions with kernel-width 3 followed by 36 4-head Transformer blocks~\cite{vaswani2017attention} with self-attention dimension $D_{tr} = 768$. The 2nd, 4th and the final convolutions in the front-end have stride 2, so the overall sub-sampling rate of the model is 8. The AMs take 80-channel log-mel filterbanks as input and are trained end-to-end with CTC loss.

\subsubsection{Language Model}
For fair comparison with existing works, IPL experiments in Table \ref{tab:comparison} use BS decoding with the 4-gram LM (200k top words) used in~\cite{synnaeve2019end}, which is trained on the official \librispeech{} LM corpus with transcriptions in \librivox{} excluded, denoted as $LS \setminus LV$. The same LM is used for the final beam-search decoding of trained models. Also we are using Transfomer LM from~\cite{synnaeve2019end} for the final beams rescoring. For IPL experiments, where beam rescoring is used for PL generation in addition to the $n$-gram BS decoding, transfomer LM (with the same architecture as transformer LM from~\cite{synnaeve2019end}) is trained on $LS \setminus LV$. As an ablation study, we train 5-gram LMs on $GB \setminus LV \setminus LS$ and $GB \setminus LV$ with top 200k words in each and without pruning using the KenLM toolkit~\cite{heafield2011kenlm}. The LMs perplexities are listed in Table~\ref{tab:lm}.
\begin{table}[h!]
\caption{Perplexities of language models.\label{tab:lm}}
\centering
\setlength\tabcolsep{4pt} 
\begin{tabular}{cccccc}
    \toprule
        Data & $LS \setminus LV$ & $GB \setminus LV \setminus LS$ & $GB \setminus LV$ & Transf. $LS \setminus LV$ & Transf. \\
    \midrule
        dev-clean & 161.7 & 101.6 & 99.7 & 58.3 & 48.2 \\
        dev-other & 152.5 & 112.9 & 110.5 & 59.3 & 50.2 \\
    \bottomrule
\end{tabular}
\end{table}


\subsection{Model Training}
\label{sec:training}
We use word pieces (WP)~\cite{kudo2018sentencepiece} as modeling units in our experiments. Following~\cite{synnaeve2019end}, we use the same 10k WP estimated from the training transcriptions, if full \librispeech{} training set is used. If \tco{} is in use, we switch to the 5k WP estimated on \tco{} transcriptions as in~\cite{hsu2020semi}. We use a lexicon, including words only in training and validation sets, to limit the search space of the BS decoding in IPL. For \llten{} setup, the same units and lexicon as \tco{} are used.

Dropout~\cite{srivastava2014dropout} and layer drop~\cite{fan2019reducing} are tuned and used to regularize each model. For models that either use \llten{} as labeled data or trained without \librivox{}, we set both dropout and layer drop to 0.3; for models trained on \librivox{}, layer drop is set to 0.2 while dropout is 0.2 and 0.15 for models using 100 and 960 hours labeled data, respectively. All models are trained on 64 GPUs with a batch size of 4 per GPU if using \librivox{} and 6 otherwise. We use the Adagrad~\cite{duchi2011adaptive} optimizer; the learning rate is initialized to 0.03 and is never decreased for models trained on \librivox{} but is halved once at epoch 800 for \librispeech{}-only models. 

In terms of IPL training, we implemented the automated pipeline in wav2letter++~\cite{pratap2018wav2letter}. For models trained with \librivox{} data, we use only BS with \ngrams{} LM in decoding; while for models trained on \librispeech{} only, we apply two stage decoding in PL generating: 1) BS decoding with \ngrams{} LM and 2) beam rescoring with Transformer LM. We use random search with 256 jobs to optimize the hyper-parameters in decoding~\cite{zeghidour2018fully} on \devother{} and use the optimal values in the subsequent pseudo-labeling. As mentioned in Section~\ref{sec:px_estimation}, we only select 20\% to 40\% of the data in each round of PL if \librivox{} is the unlabeled dataset. Otherwise, if the rest of \librispeech{} is the unlabeled set, the entire unlabeled set is pseudo-labeled. Pseudo-labels are regenerated every 10 epochs. Note that there is no filtering applied to the pseudo-labels generated, i.e. the whole unlabeled set will be used in training.

\subsection{Results}
\label{sec:training_strategy}

In this section we compare our results with other recent work in semi-supervised learning. All results are listed in Table~\ref{tab:comparison}. 
Given \llten{} as labeled data, our method reaches 26.02\% and 19.92\% on \testother{} with the full \librispeech{} or \librivox{} as unlabeled data, respectively; while with \tco{}, we further get 8.95\% and 7.11\% on \testother{} with the rest of \librispeech{} or \librivox{} as unlabeled data, respectively. 
If all of \librispeech{} is used as labeled data, we achieve 4.01\% on \testother{}. Our result achieves a clear state-of-the-art in all the three semi-supervised learning setups.\footnote{The results of this work were first published in May 2020 and the works that we are comparing to here are selected by then. Comparison with the latest methods is shown in Appendix.} The final decoding and rescoring strategies are the same as in \cite{synnaeve2019end}.

\begin{table*}[t!]
\caption{
Comparison of WER with other semi-supervised methods on \librispeech{} (LS) and \librivox{} (LV) data. The beam-search decoding with 4-gram $LS \setminus LV$ LM is used in IPL training for pseudo-labels generation (for models used \librispeech{} as unlabeled data beam rescoring with transformer $LS \setminus LV$ LM is applied for pseudo-labels generation). LM column refers either to the greedy path ("-") or to the final decoding and rescoring ("$^*$") with external~LMs.\label{tab:comparison}}
\begin{center}
\setlength\tabcolsep{5pt} 
\begin{tabular}{@{}cccccccc@{}}
\toprule
\multirow{2}{*}{Method} & \multicolumn{2}{c}{Data (hours)} & \multirow{2}{*}{LM} & \multicolumn{2}{c}{Dev WER} & \multicolumn{2}{c}{Test WER} \\
\cmidrule(lr){2-3} \cmidrule(lr){5-6} \cmidrule(lr){7-8}
                        & Labeled    & Unlabeled     &                     & clean        & other        & clean         & other        \\
\midrule
Semi-supervision with PL~\cite{kahn2019self} & LS-100 & LS-360 & GCNN & 5.37 & 22.13 & 5.93 & 24.07 \\
Local Prior Matching~\cite{hsu2020semi} & LS-100 & LS-860 & GCNN & 4.87 & 13.84 & 4.88 & 15.28 \\
DeCoAR~\cite{ling2020deep} & LS-100 & LS-860 & 4-gram & - & - & 4.74 & 12.20 \\
vq-wav2vec + BERT~\cite{baevski2019effectiveness} & LS-100 & LS-860 & 4-gram & 4.0 & 10.9 & 4.5 & 12.1 \\
Semi-supervision with PL, CTC ~\cite{synnaeve2019end} & \multirow{2}{*}{LS-960} & \multirow{2}{*}{LV-54K} & GCNN + Transf.$^*$ & 2.01 & 3.95 & 2.31 & 4.54 \\
Semi-supervision with PL, S2S ~\cite{synnaeve2019end} & & & GCNN WP + Transf.$^*$ & 2.00 & 3.65 & 2.09 & 4.11 \\
\midrule
\multirow{10}{*}{Ours, IPL} & \multirow{4}{*}{LL-10} & \multirow{2}{*}{LS-960} & - & 23.84 & 25.70 & 24.58 & 26.44 \\
 & & & 4-gram + Transf.$^*$ & 23.51 & 25.48 & 24.37 & 26.02  \\
 \cmidrule(lr){3-8}
 & & \multirow{2}{*}{LV-54K} & - & 19.76 & 21.67 & 20.63 & 22.38  \\
 & & & 4-gram + Transf.$^*$ & 17.00 & 19.34 & 18.03 & 19.92  \\
 \cmidrule(lr){2-8}
 & \multirow{4}{*}{LS-100} & \multirow{2}{*}{LS-860} & - & 5.41 & 9.32 & 5.95 & 10.28 \\
 & & & 4-gram + Transf.$^*$ & 4.98 & 7.97 & 5.59 & 8.95 \\
 \cmidrule(lr){3-8}
 & & \multirow{2}{*}{LV-54K} & - & 4.35 & 7.90 & 5.07 & 8.84 \\
 & & & 4-gram + Transf.$^*$ & 3.19 & 6.14 & 3.72 & 7.11 \\
 \cmidrule(lr){2-8}
 & \multirow{2}{*}{LS-960} & \multirow{2}{*}{LV-54K} & - & 2.05 & 4.12 & 2.21 & 4.71 \\
 & & & 4-gram + Transf.$^*$ & 1.85 & 3.26 & 2.10 & 4.01 \\
\bottomrule
\end{tabular}
\end{center}
\end{table*}
\label{sec:acoustic_models}

\subsection{Analysis}

\subsubsection{Effectiveness}
We conduct 3 rounds of pseudo-labeling on the entire unlabeled set and retraining a new AM from scratch. All the models in this section are trained with \ngrams{} BS decoding in pseudo-labeling. As shown in Table \ref{tab:effectiveness}, WER on \devother{} decreases with better PL generated, but the marginal gain diminishes as iterations continue. IPL, however, clearly outperforms the 3 rounds PL baseline, indicating it is effective to accumulate gains through training with more (up to 80) rounds of PL updates. 

\begin{table}[h!]
\caption{WER of greedy path on \devother{} for IPL and training from scratch for multiple rounds. 4-gram $LS \setminus LV$ LM is used for pseudo-labels generation. \label{tab:effectiveness}}
\centering
\begin{tabular}{ccccccc}
    \toprule
        \multicolumn{2}{c}{Data} & \multicolumn{4}{c}{\# Rounds of PL} & \multirow{2}{*}{IPL} \\ 
        \cmidrule(lr){1-2} \cmidrule(lr){3-6}
        Labeled  & Unlabeled & 0 & 1     & 2    & 3    &       \\
    \midrule
        LS-100  & LS-860  & 27.76 & 17.1 & 15.8 & 15.09 & 10.69 \\
        LS-100  & LS + LV   & 27.76 & 16.3 & 12.9 & 10.95 & 7.90 \\
        LS-960 & LV-54K  & 7.31 & 5.00 & 4.69 & 4.57 & 4.12  \\
    \bottomrule
\end{tabular}
\end{table}

\subsubsection{Efficiency}
Given the same amount of time for 3 rounds of PL training in Table \ref{tab:effectiveness} to finish, which is 4, 11 and 17 days from top to bottom, IPL achieves WER 10.69\%, 8.50\% and 4.14\%, respectively. To achieve the same WER after round 3, however, IPL takes only 0.7, 3.3 and 8 days. This efficiency derives directly from the two proposed changes in IPL: 1) fine tuning the existing model with new labels to save computation in re-bootstrapping and 2) down sampling the unlabeled set to shorten the PL time, i.e. 20\% down-sample rate leads to 5 time speed up in labeling.

\subsubsection{LM Study}
Comparison of IPL with different LMs is shown in Table~\ref{tab:IPL_LM} with LMs perplexity in Table~\ref{tab:lm}. All the models in this section are trained with \ngrams{} BS decoding in pseudo-labeling. Given a better LM, IPL better transfers LM knowledge into AM and achieves better performance. IPL can thus effectively leverage large amount of unpaired text, in addition to unpaired audio. However, although the difference in perplexity between the $GB \setminus LV$ and $GB \setminus LV \setminus LS$ LMs is small, there is still a large gap in WER. This is because the LM implicitly leaks the labels of unlabeled audio. Fortunately, comparing WER between $LS \setminus LV$ and $GB \setminus LV \setminus LS$, when the transcription leaking is completely removed, it is still possible to reach similar (or even better) WER by utilizing more in-domain text.

\begin{table}[h!]
\caption{IPL training with different LMs in BS decoding. WER on \devother{} (\testother{}) is reported for the greedy path (top), an extra BS decoding with the same \ngrams{} LM used in IPL training (middle) and Transformer LM rescoring (bottom). \label{tab:IPL_LM}}
\centering
\setlength\thickmuskip{0mu}
\setlength\thinmuskip{0mu}
\setlength\medmuskip{0mu}
\begin{tabular}{cccccc}
    \toprule
        \multirow{2}{*}{Decoding} & \multicolumn{2}{c}{Data} & \multicolumn{3}{c}{Language Model}  \\ 
        \cmidrule(lr){2-3} \cmidrule(lr){4-6}
        & Labeled  & Unlabeled & $LS \setminus LV$ & $GB \setminus LV \setminus LS$ & $GB \setminus LV$       \\
    \midrule
        \multirow{3}{*}{None} & LS-100 & LS-860  & 10.69 (11.48) & 10.80 (11.61) & 10.19 (11.09) \\
        & LS-100  & LS + LV   & 7.90 (8.84) & 7.21 (8.28) & 6.82 (7.89) \\
        & LS-960 & LV-54K  & 4.12 (4.71) & - & 4.02 (4.42)  \\
    \cmidrule(lr){1-6}
        \multirow{3}{*}{4-gram} & LS-100  & LS-860  & 10.05 (10.90) & 9.82 (10.49) & 9.09 (9.82) \\
        & LS-100  & LS + LV   & 7.21 (8.19) & 6.70 (7.74) & 6.15 (7.35) \\
        & LS-960 & LV-54K  & 3.67 (4.33) & - & 3.65 (4.10)  \\
    \cmidrule(lr){1-6}
        \multirow{3}{*}{Trans.} & LS-100  & LS-860  & 8.72 (9.51) & 8.90 (9.67) & 8.25 (9.11) \\
        & LS-100  & LS + LV   & 6.14 (7.11)  & 5.96 (6.99) & 5.56 (6.71) \\
        & LS-960 & LV-54K  & 3.26 (4.01) & - & 3.42 (3.83) \\
    \bottomrule
\end{tabular}
\end{table}

\subsubsection{Decoding Parameters Mismatch}
As shown in Table~\ref{tab:IPL_LM}, there is still an observable improvement in WER across greedy path and the BS decoding with \ngrams{} LM. One inhibitor to transferring LM knowledge into the AM is a mismatch in decoding parameters: the parameters optimized on \devother{} may not be optimal for decoding unlabeled audio. Thus, in the final stage of IPL training, the marginal WER improvement on \devother{} is not reflected similarly on the one on unlabeled audio, so as to prevent the AM from improving.

\section{Conclusion}
We have shown that iterative pseudo-labeling can give superior results in both standard and low-resource settings and provides an efficient algorithm with which to train as compared to conventional pseudo-labeling approaches. Iterative pseudo-labeling benefits from beam-search decoding with a language model and data augmentation along with dataset sub-sampling which also improves efficiency. With our Transformer acoustic model, IPL achieves the state-of-the art results on \librispeech{} test sets. 


\clearpage
\newpage


\clearpage
\newpage
\appendix

\section{Learning Curves}
\subsection{Typical IPL curves}
We provide training curves for three scenarios with different amount of labeled data. There is an obvious inflection point when IPL starts and unlabeled data is in use. In the first several rounds of IPL, WER decreases dramatically each time when new data and their PLs are generated and consumed. IPL is also able to keep accumulating marginal gains until the end of training. Surprisingly, IPL is able to bootstrap from an immature model, even with about 80\% WER. Key to the success here is the LM and lexicon that limits the search space of words. 

\begin{figure}[!h]
    \centering
    \input{figures/a1_ipl.tikz}
    \caption{Typical IPL training curves on \devother{}. The labeled training data used in each sub-plot is (left) full \librispeech{}, (middle) \tco{}, (right) \llten{}. \librispeech{} and \librivox{} are used as the unlabeled data. Each epoch may contain different amount of data depending on the subset size selected from unlabeled data.}
\label{fig:ipl_curve}
\end{figure}
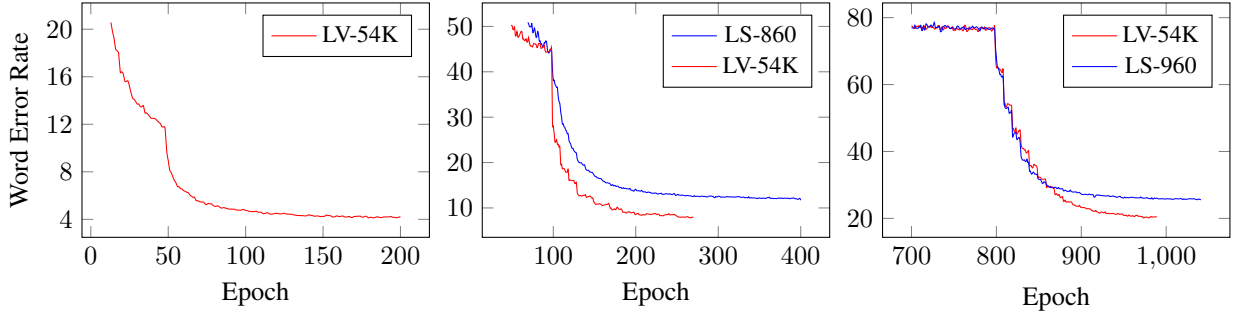

\subsection{IPL v.s. Training From Scratch}
If the AM is trained from scratch each time as the PL evolves, we can see the gain is diminishing. In addition, both training AM from scratch and label the whole 54K hours dataset are time consuming. Ignoring the labeling time and given the same amount of training time for an AM to converge from scratch, IPL can always outperform and reach better WER. 
\begin{figure}[!h]
    \centering
    \input{figures/a2_ipl_vs_scratch.tikz}
    \caption{Comparison between IPL and training from scratch with pseudo-labels. Label and unlabeled training data used in these experiments are \tco{} and \librivox{} plus the rest of \librispeech{}, respectively.}
\label{fig:ipl_vs_scratch}
\end{figure}
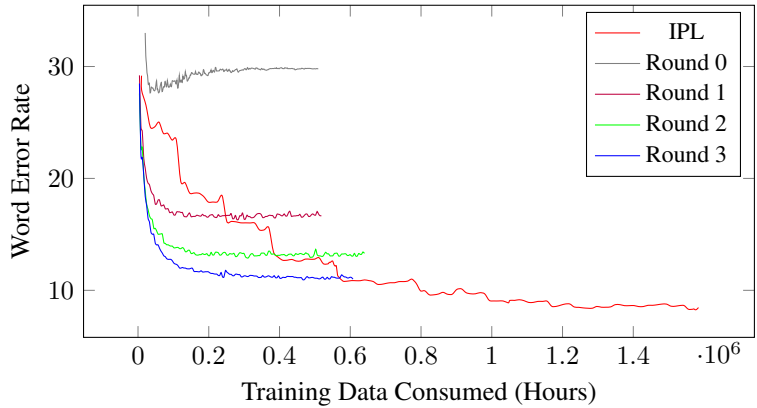

\subsection{IPL v.s. Retraining}
\begin{figure}[!h]
    \centering
    \input{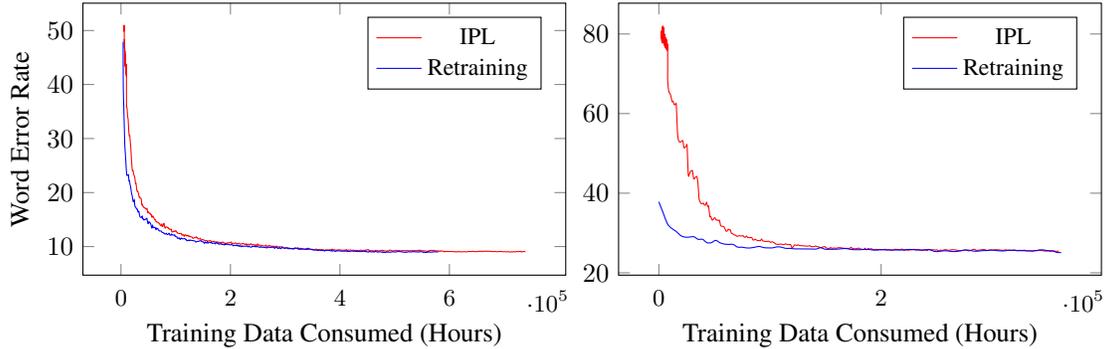}
    \caption{Retraining a final model from scratch with the pseudo-labels generated from a fully converged IPL model. Labeled training data used in the subplots are (left) \tco{} and (right) \llten{}, while the rest of \librispeech{} is used as unlabeled data in both.}
\label{fig:ipl_ft_final}
\end{figure}

We show in Section 1 that fine tuning and training from scratch converge to the same ball park in the early stage of IPL. This experiment is designed to verify if this still holds when the IPL model is converged. As shown in Figure \ref{fig:ipl_ft_final}, if we train a new AM from scratch using the pseudo-labels generated from a fully converged IPL model, it will not outperform the IPL model. Even with beam search decoding, both models can still reach similar results. For the model trained on \tco{}, WER are 8.83 and 8.87 for IPL and retraining models, respectively; while for models trained on \llten{}, WERs are 25.69 and 25.62. 

\section{Decoding Parameters}
\subsection{LM weights for different LMs}
We conduct grid search over LM weight used in beam-search decoding and rescoring for an IPL model. Specifically, we use a 4-gram LM in the beam-search decoding to generate a beam of candidates and rescore the beam using an NN-LM. The final WER is not sensitive to the LM weight used in the beam-search decoding, which means as long as the LM weight is reasonable (e.g. between 0 to 2) the good candidates can always be preserved in the beam and later selected out by rescoring. This enables us to pay less attention to the beam-search decoder parameter tuning and focus on rescoring. 

\begin{figure}[!h]
    \centering
    \vspace{-0.8cm}
    \includegraphics[width=.5\linewidth]{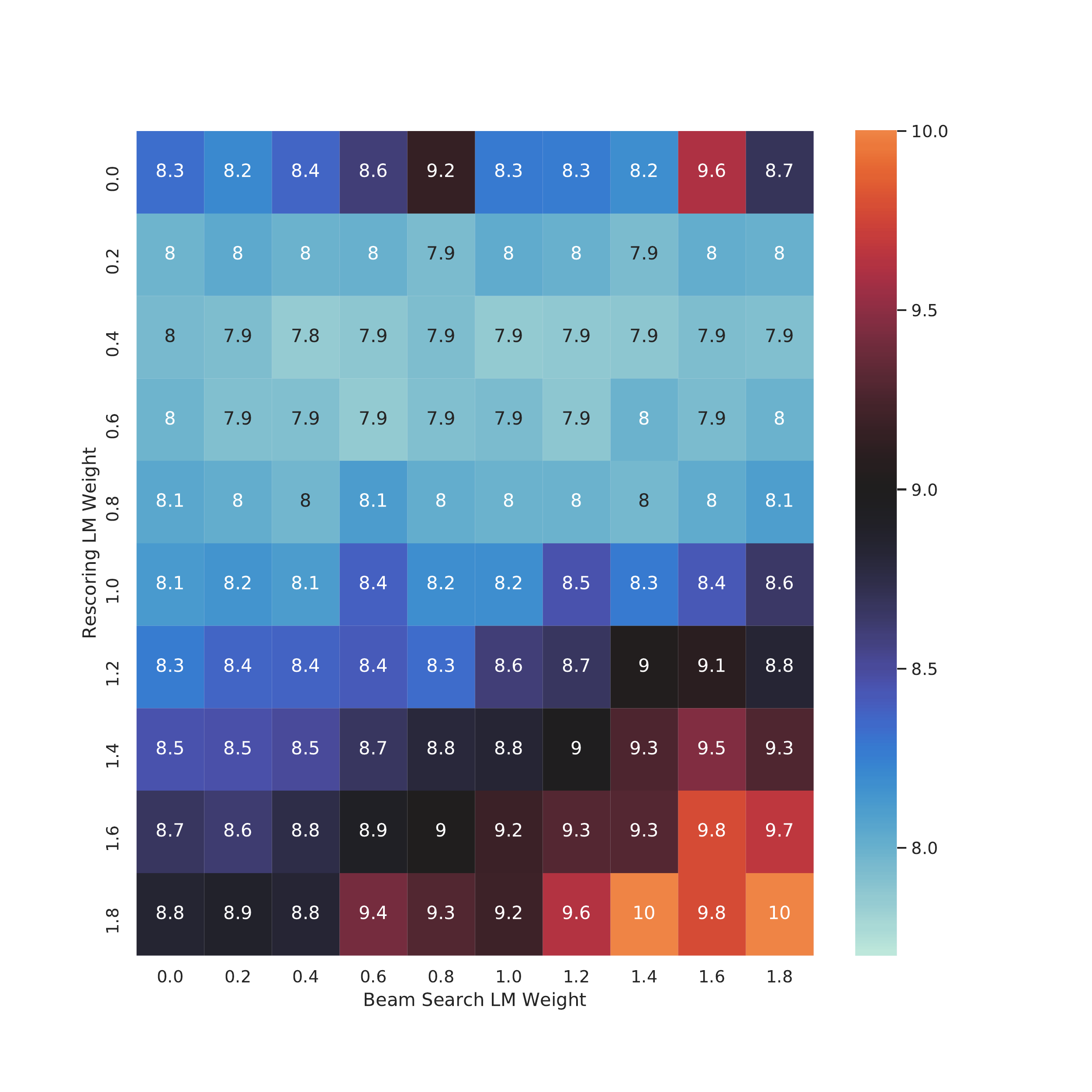}
    \vspace{-0.8cm}
    \caption{WER with different LM weights used in beam-search decoding and rescoring.}
\label{fig:gain}
\end{figure}

\subsection{LM weights on different dataset}
\begin{figure}[!h]
    \centering
    \vspace{-0.8cm}
    \includegraphics[width=.5\linewidth]{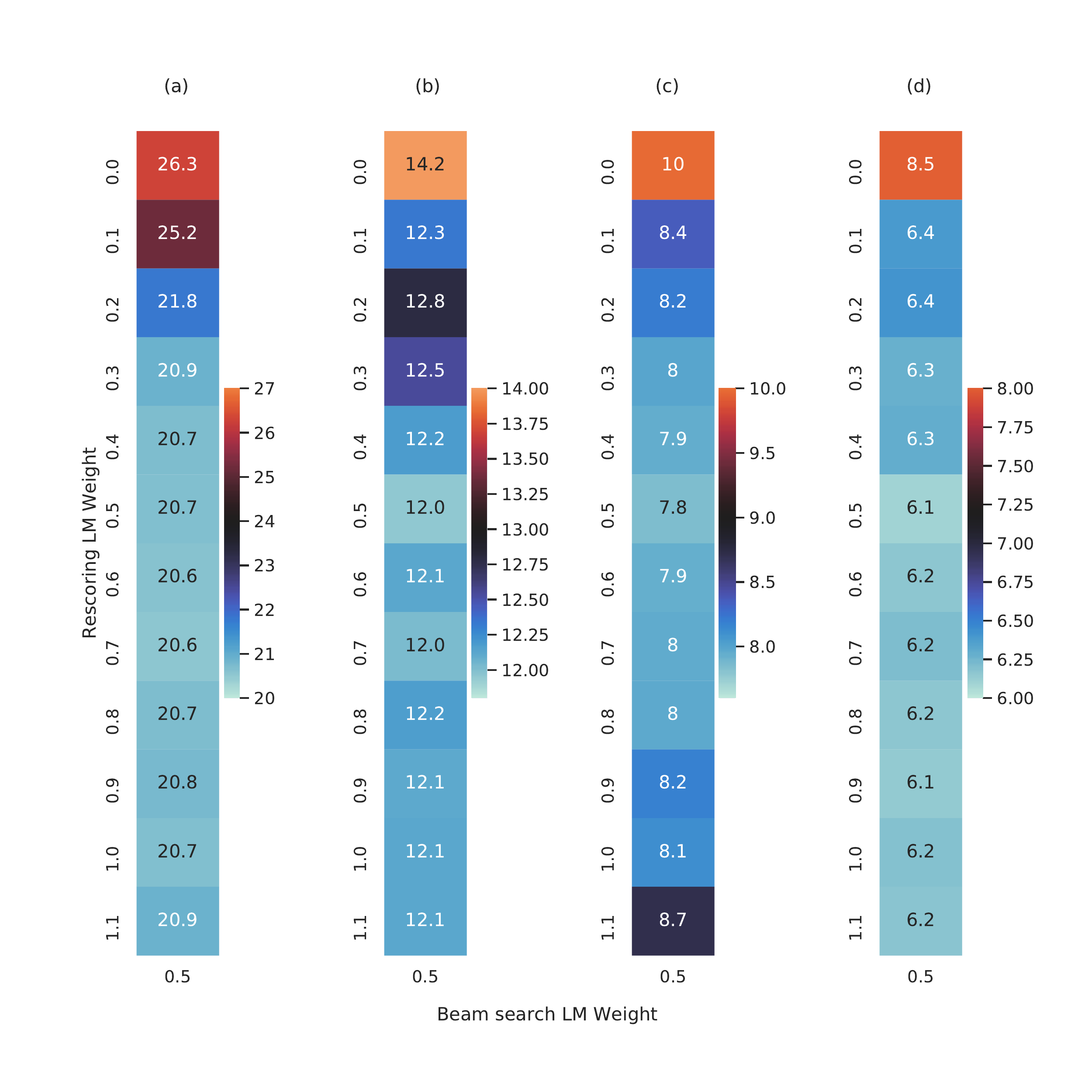}
    \vspace{-0.6cm}
    \caption{Comparison of the WER heatmap of rescoring LM weight on development and training set. Left: decoding a checkpoint of an AM that is not fully converged, with (a) WER on \devother{} and (b) WER on \tco{} and \tof{}. Right: decoding a fully converged AM, with (c) WER on \devother{} and (d) WER on \tco{} and \tof{}. }
\label{fig:search_dev_train}
\end{figure}

We conduct parameter sweep on both development and training set to see if there is a mismatch between the optimal decoding/rescoring parameters on them. As shown in Figure \ref{fig:search_dev_train}, the optimal LM weight is aligned between development and train for both immature and converged model. This shows that the development set can be used as a good proxy of decoder parameter tuning.  

\section{Comparison With The Latest Methods}

\begin{table*}[b!]
\caption{
Comparison of WER with lastest semi-supervised methods on \librispeech{} (LS) and \librivox{} (LV) data. 
\label{tab:comparison_latest}}
\begin{center}
\setlength\tabcolsep{5pt} 
\begin{tabular}{@{}cccccccc@{}}
\toprule
\multirow{2}{*}{Method} & \multicolumn{2}{c}{Data (hours)} & \multirow{2}{*}{LM} & \multicolumn{2}{c}{Dev WER} & \multicolumn{2}{c}{Test WER} \\
\cmidrule(lr){2-3} \cmidrule(lr){5-6} \cmidrule(lr){7-8}
                        & Labeled    & Unlabeled     &                     & clean        & other        & clean         & other        \\
\midrule
\multirow{2}{*}{Improved T/S~\cite{park2020improved}} & LS-100 & LS-860 & LSTM & 3.9 & 8.8 & 4.2 & 8.6 \\
 & LS-960 & LV-54K & LSTM & 1.6 & 3.4 & 1.7 & 3.4 \\
\midrule
\multirow{5}{*}{wav2vec~2.0~\cite{baevski2020wav2vec}} & LL-10 & LS-860 & Transf. & 2.9 & 5.7 & 3.2 & 6.1 \\
  & LL-10 & LV-54K & Transf. & 2.5 & 5.2 & 2.6 & 5.2 \\
  & LS-100 & LS-860 & Transf. & 2.1 & 4.8 & 2.3 & 5.0 \\
  & LS-100 & LV-54K & Transf. & 2.0 & 4.1 & 2.1 & 4.4 \\
  & LS-960 & LV-54K & Transf. & 1.6 & 3.2 & 1.9 & 3.5 \\
\midrule
\multirow{10}{*}{Ours, IPL} & \multirow{4}{*}{LL-10} & \multirow{2}{*}{LS-960} & - & 23.84 & 25.70 & 24.58 & 26.44 \\
 & & & 4-gram + Transf.$^*$ & 23.51 & 25.48 & 24.37 & 26.02  \\
 \cmidrule(lr){3-8}
 & & \multirow{2}{*}{LV-54K} & - & 19.76 & 21.67 & 20.63 & 22.38  \\
 & & & 4-gram + Transf.$^*$ & 17.00 & 19.34 & 18.03 & 19.92  \\
 \cmidrule(lr){2-8}
 & \multirow{4}{*}{LS-100} & \multirow{2}{*}{LS-860} & - & 5.48 & 9.32 & 5.95 & 10.31 \\
 & & & 4-gram + Transf.$^*$ & 4.98 & 7.97 & 5.59 & 8.95 \\
 \cmidrule(lr){3-8}
 & & \multirow{2}{*}{LV-54K} & - & 4.35 & 7.90 & 5.07 & 8.84 \\
 & &  & 4-gram + Transf.$^*$ & 3.19 & 6.14 & 3.72 & 7.11 \\
 \cmidrule(lr){2-8}
 & \multirow{2}{*}{LS-960} & \multirow{2}{*}{LV-54K} & - & 2.05 & 4.12 & 2.21 & 4.71 \\
 & & & 4-gram + Transf.$^*$ & 1.85 & 3.26 & 2.10 & 4.01 \\
\bottomrule
\end{tabular}
\end{center}
\end{table*}

In Table \ref{tab:comparison_latest}, we mainly compare with two latest works: one with multi-rounds of pseudo-labeling \cite{park2020improved} and the other with unsupervised pre-training~\cite{baevski2020wav2vec}. Both works achieve better results than us but with different approach. \cite{park2020improved} uses the same pseudo-labeling, but is equipped with 1) Listen, Attend and Spell (LAS) network as an acoustic model reaching better WER on labeled data only and 2) filtering mechanism on the pseudo-labels. \cite{baevski2020wav2vec} learns pre-trained speech features on large scale clean speech dataset and further fine tunes the model on labeled transcriptions only.

IPL, however, can easily absorb and combine the advantages of both methods by leveraging a better acoustic model and pre-trained features. This will be a future work to further push the performance of semi-supervised ASR.

\end{document}

%% file: figures/motivation.tikz
\begin{tikzpicture}
    \begin{axis}[
        width = 7.2cm, height = 4.2cm,
        ytick={20,35,50},
        xtick={0, 20000,40000,60000,80000},
        ylabel={Word Error Rate},
        xlabel={Training Data Consumed (hours)},
        every x tick scale label/.style={at={(xticklabel cs:1)},anchor=south west},
    ]
        \addplot[blue, smooth] coordinates {
            (1713.7699999999995, 49.13)
            (1814.5799999999995, 48.2)
            (1915.3899999999994, 46.91)
            (2016.1999999999994, 46.91)
            (2117.0099999999993, 46.57)
            (2217.8199999999993, 44.58)
            (2318.629999999999, 44.58)
            (2419.439999999999, 45.46)
            (2520.249999999999, 44.58)
            (2621.059999999999, 43.05)
            (2721.869999999999, 43.05)
            (2822.679999999999, 42.02)
            (2923.489999999999, 40.77)
            (3024.299999999999, 40.59)
            (3125.1099999999988, 39.69)
            (3225.9199999999987, 39.13)
            (3326.7299999999987, 38.77)
            (3427.5399999999986, 38.31)
            (3528.3499999999985, 38.31)
            (3629.1599999999985, 37.8)
            (3729.9699999999984, 36.65)
            (3830.7799999999984, 36.65)
            (3931.5899999999983, 37.71)
            (4032.3999999999983, 36.5)
            (4133.209999999998, 36.5)
            (4234.019999999999, 37.12)
            (4334.829999999999, 35.68)
            (4435.639999999999, 35.6)
            (4536.45, 35.3)
            (4637.26, 35.1)
            (4738.070000000001, 34.61)
            (4838.880000000001, 34.61)
            (4939.690000000001, 34.59)
            (5040.500000000002, 34.59)
            (5141.310000000002, 34.27)
            (5242.120000000003, 34.27)
            (5342.930000000003, 34.44)
            (5443.740000000003, 34.34)
            (5544.550000000004, 33.96)
            (5645.360000000004, 33.96)
            (5746.170000000005, 34.09)
            (5846.980000000005, 33.78)
            (5947.790000000005, 33.78)
            (6048.600000000006, 33.91)
            (6149.410000000006, 33.07)
            (6250.220000000007, 33.07)
            (6351.030000000007, 33.02)
            (6451.840000000007, 32.25)
            (6552.650000000008, 32.25)
            (6653.460000000008, 33.23)
            (6754.270000000009, 33.1)
            (6855.080000000009, 32.85)
            (6955.890000000009, 32.85)
            (7056.70000000001, 32.25)
            (7157.51000000001, 32.25)
            (7258.320000000011, 32.7)
            (7359.130000000011, 32.7)
            (7459.940000000011, 31.86)
            (7560.750000000012, 31.86)
            (7661.560000000012, 31.98)
            (7762.370000000013, 31.98)
            (7863.180000000013, 32.35)
            (7963.990000000013, 32.33)
            (8064.800000000014, 32.33)
            (8165.610000000014, 31.96)
            (8266.420000000015, 31.96)
            (8367.230000000014, 31.99)
            (8468.040000000014, 32.02)
            (8568.850000000013, 32.02)
            (8669.660000000013, 32.34)
            (8770.470000000012, 32.12)
            (8871.280000000012, 32.12)
            (8972.090000000011, 31.56)
            (9072.90000000001, 31.56)
            (9173.71000000001, 32.12)
            (9274.52000000001, 31.7)
            (9375.330000000009, 31.25)
            (9476.140000000009, 31.25)
            (9576.950000000008, 31.85)
            (9677.760000000007, 31.85)
            (9778.570000000007, 32.17)
            (9879.380000000006, 32.12)
            (9980.190000000006, 31.92)
            (10081.000000000005, 31.92)
            (10181.810000000005, 33.05)
            (10282.620000000004, 31.68)
            (10383.430000000004, 31.68)
            (10484.240000000003, 31.19)
            (10585.050000000003, 31.19)
            (10685.860000000002, 31.55)
            (10786.670000000002, 31.55)
            (10887.480000000001, 32.02)
            (10988.29, 31.53)
            (11089.1, 31.53)
            (11189.91, 32.32)
            (11290.72, 32.32)
            (11391.529999999999, 31.77)
            (11492.339999999998, 31.47)
            (11593.149999999998, 31.47)
            (11693.959999999997, 31.67)
            (11794.769999999997, 31.23)
            (11895.579999999996, 31.23)
            (11996.389999999996, 27.64)
            (12097.199999999995, 26.71)
            (12560.889999999996, 26.71)
            (13024.579999999996, 26.84)
            (13488.269999999997, 25.69)
            (13951.959999999997, 25.69)
            (14415.649999999998, 25.04)
            (14879.339999999998, 24.77)
            (15343.029999999999, 24.77)
            (15806.72, 24.72)
            (16270.41, 24.72)
            (16734.1, 24.51)
            (17197.789999999997, 24.51)
            (17661.479999999996, 24.18)
            (18125.169999999995, 24.18)
            (18588.859999999993, 24.2)
            (19052.549999999992, 24.2)
            (19516.23999999999, 24.0)
            (19979.92999999999, 24.0)
            (20443.619999999988, 24.1)
            (20907.309999999987, 23.81)
            (21370.999999999985, 23.81)
            (21834.689999999984, 23.86)
            (22298.379999999983, 23.51)
            (22762.06999999998, 23.51)
            (23225.75999999998, 23.32)
            (23689.44999999998, 23.32)
            (24153.139999999978, 23.84)
            (24616.829999999976, 23.84)
            (25080.519999999975, 23.77)
            (25544.209999999974, 23.69)
            (26007.899999999972, 23.68)
            (26471.58999999997, 23.68)
            (26935.27999999997, 23.35)
            (27398.96999999997, 23.35)
            (27862.659999999967, 23.06)
            (28326.349999999966, 23.06)
            (28790.039999999964, 23.11)
            (29253.729999999963, 23.11)
            (29717.419999999962, 23.27)
            (30181.10999999996, 23.21)
            (30644.79999999996, 23.21)
            (31108.489999999958, 23.12)
            (31572.179999999957, 23.12)
            (32035.869999999955, 23.25)
            (32499.559999999954, 23.36)
            (32963.249999999956, 23.36)
            (33426.93999999996, 23.32)
            (33890.62999999996, 23.32)
            (34354.31999999996, 23.02)
            (34818.009999999966, 23.02)
            (35281.69999999997, 23.52)
            (35745.38999999997, 23.43)
            (36209.07999999997, 23.43)
            (36672.769999999975, 23.45)
            (37136.45999999998, 23.54)
            (37600.14999999998, 23.41)
            (38063.83999999998, 23.41)
            (38527.529999999984, 23.12)
            (38991.21999999999, 23.12)
            (39454.90999999999, 23.14)
            (39918.59999999999, 23.54)
            (40382.28999999999, 23.17)
            (40845.979999999996, 23.17)
            (41309.67, 23.41)
            (41773.36, 23.22)
            (42237.05, 23.13)
            (42700.740000000005, 22.81)
            (43164.43000000001, 22.81)
            (43628.12000000001, 22.9)
            (44091.81000000001, 22.9)
            (44555.500000000015, 22.75)
            (45019.19000000002, 22.75)
            (45482.88000000002, 23.03)
            (45946.57000000002, 23.05)
            (46410.260000000024, 23.05)
            (46873.950000000026, 23.05)
            (47337.64000000003, 23.34)
            (47801.33000000003, 23.15)
            (48265.02000000003, 23.15)
            (48728.710000000036, 23.27)
            (49192.40000000004, 23.05)
            (49656.09000000004, 23.05)
            (50119.78000000004, 22.9)
            (50583.470000000045, 22.9)
            (51047.16000000005, 23.04)
            (51510.85000000005, 22.8)
            (51974.54000000005, 22.8)
            (52438.230000000054, 23.27)
            (52901.92000000006, 22.9)
            (53365.61000000006, 22.9)
            (53829.30000000006, 22.91)
            (54292.99000000006, 22.79)
            (54756.680000000066, 22.6)
            (55220.37000000007, 22.6)
            (55684.06000000007, 22.83)
            (56147.75000000007, 22.83)
            (56611.440000000075, 22.72)
            (57075.13000000008, 22.72)
            (57538.82000000008, 23.08)
            (58002.51000000008, 22.66)
            (58466.200000000084, 22.51)
            (58929.89000000009, 22.43)
            (59393.58000000009, 22.43)
            (59857.27000000009, 22.72)
            (60320.960000000094, 23.09)
            (60784.650000000096, 22.76)
            (61248.3400000001, 22.76)
            (61712.0300000001, 22.51)
            (62175.7200000001, 22.51)
            (62639.410000000105, 22.82)
            (63103.10000000011, 22.72)
            (63566.79000000011, 22.49)
            (64030.48000000011, 22.49)
            (64494.170000000115, 22.52)
            (64957.86000000012, 22.59)
            (65421.55000000012, 22.74)
            (65885.24000000012, 23.05)
            (66348.93000000012, 22.65)
            (66812.62000000013, 22.65)
            (67276.31000000013, 22.82)
            (67740.00000000013, 22.82)
            (68203.69000000013, 22.87)
            (68667.38000000014, 22.87)
            (69131.07000000014, 22.46)
            (69594.76000000014, 22.46)
            (70058.45000000014, 23.04)
            (70522.14000000014, 22.38)
            (70985.83000000015, 22.38)
            (71449.52000000015, 22.87)
            (71913.21000000015, 22.63)
            (72376.90000000015, 22.61)
            (72840.59000000016, 22.61)
            (73304.28000000016, 22.69)
            (73767.97000000016, 22.5)
            (74231.66000000016, 22.5)
            (74695.35000000017, 22.67)
            (75159.04000000017, 22.24)
            (75622.73000000017, 22.24)
            (76086.42000000017, 22.7)
            (76550.11000000018, 22.72)
            (77013.80000000018, 22.34)
            (77477.49000000018, 22.34)
            (77941.18000000018, 22.46)
            (78404.87000000018, 22.77)
            (78868.56000000019, 23.02)
            (79332.25000000019, 22.67)
            (79795.94000000019, 22.67)
            (80259.6300000002, 22.69)
            (80723.3200000002, 22.69)
            (81187.0100000002, 23.0)
        };
        \addplot[red, smooth] coordinates {
            (927.38, 47.82)
            (1391.07, 40.46)
            (1854.76, 40.26)
            (2318.45, 35.89)
            (2782.14, 34.77)
            (3245.83, 32.46)
            (3709.52, 31.69)
            (4173.21, 29.91)
            (4636.9, 28.97)
            (5100.589999999999, 28.38)
            (5564.279999999999, 28.38)
            (6027.969999999998, 27.92)
            (6491.659999999998, 27.92)
            (6955.349999999998, 27.66)
            (7419.039999999997, 27.35)
            (7882.729999999997, 26.94)
            (8346.419999999996, 26.94)
            (8810.109999999997, 27.49)
            (9273.799999999997, 26.08)
            (9737.489999999998, 25.89)
            (10201.179999999998, 25.89)
            (10664.869999999999, 25.42)
            (11128.56, 25.42)
            (11592.25, 25.05)
            (12055.94, 24.7)
            (12519.630000000001, 24.67)
            (12983.320000000002, 23.96)
            (13447.010000000002, 23.96)
            (13910.700000000003, 24.21)
            (14374.390000000003, 24.21)
            (14838.080000000004, 24.58)
            (15301.770000000004, 23.86)
            (15765.460000000005, 23.23)
            (16229.150000000005, 23.23)
            (16692.840000000004, 23.81)
            (17156.530000000002, 23.81)
            (17620.22, 23.61)
            (18083.91, 23.61)
            (18547.6, 23.36)
            (19011.289999999997, 23.36)
            (19474.979999999996, 23.32)
            (19938.669999999995, 23.32)
            (20402.359999999993, 23.08)
            (20866.049999999992, 23.08)
            (21329.73999999999, 23.32)
            (21793.42999999999, 23.35)
            (22257.119999999988, 23.0)
            (22720.809999999987, 23.0)
            (23184.499999999985, 23.17)
            (23648.189999999984, 22.87)
            (24111.879999999983, 22.84)
            (24575.56999999998, 22.84)
            (25039.25999999998, 22.89)
            (25502.94999999998, 22.91)
            (25966.639999999978, 22.91)
            (26430.329999999976, 22.75)
            (26894.019999999975, 22.69)
            (27357.709999999974, 22.62)
            (27821.399999999972, 22.62)
            (28285.08999999997, 22.55)
            (28748.77999999997, 22.39)
            (29212.46999999997, 22.39)
            (29676.159999999967, 22.57)
            (30139.849999999966, 22.44)
            (30603.539999999964, 22.44)
            (31067.229999999963, 22.6)
            (31530.919999999962, 22.65)
            (31994.60999999996, 22.65)
            (32458.29999999996, 22.25)
            (32921.98999999996, 22.25)
            (33385.679999999964, 22.39)
            (33849.369999999966, 22.39)
            (34313.05999999997, 22.46)
            (34776.74999999997, 22.35)
            (35240.43999999997, 22.31)
            (35704.129999999976, 22.31)
            (36167.81999999998, 22.39)
            (36631.50999999998, 22.32)
            (37095.19999999998, 22.32)
            (37558.889999999985, 22.6)
            (38022.57999999999, 22.34)
            (38486.26999999999, 22.34)
            (38949.95999999999, 22.41)
            (39413.649999999994, 22.21)
            (39877.34, 22.21)
            (40341.03, 22.1)
            (40804.72, 22.1)
            (41268.41, 22.1)
            (41732.100000000006, 22.1)
            (42195.79000000001, 22.59)
            (42659.48000000001, 22.45)
            (43123.17000000001, 22.18)
            (43586.860000000015, 22.02)
            (44050.55000000002, 22.02)
            (44514.24000000002, 22.43)
            (44977.93000000002, 22.12)
            (45441.620000000024, 22.12)
            (45905.31000000003, 22.51)
            (46369.00000000003, 22.34)
            (46832.69000000003, 22.34)
            (47296.380000000034, 22.14)
            (47760.070000000036, 22.14)
            (48223.76000000004, 22.25)
            (48687.45000000004, 22.56)
            (49151.14000000004, 22.43)
            (49614.830000000045, 22.31)
            (50078.52000000005, 22.22)
            (50542.21000000005, 22.22)
            (51005.90000000005, 22.44)
            (51469.590000000055, 22.41)
            (51933.28000000006, 22.37)
            (52396.97000000006, 22.11)
            (52860.66000000006, 22.11)
            (53324.350000000064, 22.16)
            (53788.040000000066, 22.16)
            (54251.73000000007, 22.37)
            (54715.42000000007, 22.19)
            (55179.11000000007, 22.19)
            (55642.800000000076, 22.28)
            (56106.49000000008, 22.37)
            (56570.18000000008, 22.37)
            (57033.87000000008, 22.39)
            (57497.560000000085, 22.55)
            (57961.25000000009, 22.55)
            (58424.94000000009, 22.23)
            (58888.63000000009, 22.23)
            (59352.320000000094, 22.22)
            (59816.0100000001, 22.22)
            (60279.7000000001, 22.19)
            (60743.3900000001, 22.19)
            (61207.0800000001, 22.26)
            (61670.770000000106, 22.55)
            (62134.46000000011, 22.24)
            (62598.15000000011, 22.24)
            (63061.84000000011, 22.35)
            (63525.530000000115, 22.49)
            (63989.22000000012, 22.27)
            (64452.91000000012, 22.27)
            (64916.60000000012, 22.37)
            (65380.290000000125, 22.37)
            (65843.98000000013, 22.55)
            (66307.67000000013, 22.55)
            (66771.36000000013, 22.58)
            (67235.05000000013, 22.24)
            (67698.74000000014, 22.24)
            (68162.43000000014, 22.29)
            (68626.12000000014, 22.5)
            (69089.81000000014, 22.5)
            (69553.50000000015, 22.5)
            (70017.19000000015, 22.77)
            (70480.88000000015, 22.34)
            (70944.57000000015, 22.34)
            (71408.26000000015, 22.48)
            (71871.95000000016, 22.46)
            (72335.64000000016, 22.45)
            (72799.33000000016, 22.29)
            (73263.02000000016, 22.29)
            (73726.71000000017, 22.59)
            (74190.40000000017, 22.56)
            (74654.09000000017, 22.48)
            (75117.78000000017, 22.48)
        };
        \legend{{\small fine tuning}, {\small training from scratch}}
\end{axis}
\end{tikzpicture}

%% file: figures/a1_ipl.tikz
\begin{tikzpicture}
    \begin{groupplot}[group style = {group size = 3 by 1, horizontal sep = 20pt}, width =6.2cm, height = 4.7cm]
        \nextgroupplot[ 
         	xtick={0, 50, 100, 150, 200},
         	ytick={4, 8, 12, 16, 20},
         	ylabel={Word Error Rate},
            xlabel={Epoch},
            legend style={at={(0.95,0.95)},legend columns=2,font=\small}
            ]
            \addplot[red, smooth] coordinates {
                (13, 20.57)
                (14, 19.8)
                (15, 19.25)
                (16, 18.35)
                (17, 18.19)
                (18, 17.92)
                (19, 16.42)
                (20, 16.42)
                (21, 16.23)
                (22, 15.62)
                (23, 15.62)
                (24, 15.64)
                (25, 15.18)
                (26, 14.64)
                (27, 14.2)
                (28, 14.1)
                (29, 13.91)
                (30, 13.7)
                (31, 13.7)
                (32, 13.48)
                (33, 13.48)
                (34, 13.56)
                (35, 12.96)
                (36, 12.96)
                (37, 12.82)
                (38, 12.66)
                (39, 12.51)
                (40, 12.5)
                (41, 12.5)
                (42, 12.42)
                (43, 12.29)
                (44, 12.16)
                (45, 12.04)
                (46, 11.78)
                (47, 11.78)
                (48, 11.62)
                (49, 9.75)
                (50, 8.96)
                (51, 8.13)
                (52, 7.92)
                (53, 7.57)
                (54, 7.26)
                (55, 7.07)
                (56, 6.75)
                (57, 6.75)
                (58, 6.57)
                (59, 6.55)
                (60, 6.41)
                (61, 6.35)
                (62, 6.28)
                (63, 6.24)
                (64, 6.05)
                (65, 5.95)
                (66, 5.85)
                (67, 5.85)
                (68, 5.57)
                (69, 5.57)
                (70, 5.51)
                (71, 5.46)
                (72, 5.46)
                (73, 5.43)
                (74, 5.26)
                (75, 5.26)
                (76, 5.34)
                (77, 5.3)
                (78, 5.3)
                (79, 5.16)
                (80, 5.1)
                (81, 5.1)
                (82, 5.14)
                (83, 5.05)
                (84, 4.96)
                (85, 4.96)
                (86, 4.99)
                (87, 4.94)
                (88, 4.91)
                (89, 4.87)
                (90, 4.86)
                (91, 4.86)
                (92, 4.85)
                (93, 4.85)
                (94, 4.78)
                (95, 4.78)
                (96, 4.77)
                (97, 4.77)
                (98, 4.82)
                (99, 4.82)
                (100, 4.74)
                (101, 4.73)
                (102, 4.69)
                (103, 4.66)
                (104, 4.65)
                (105, 4.65)
                (106, 4.66)
                (107, 4.68)
                (108, 4.6)
                (109, 4.6)
                (110, 4.6)
                (111, 4.49)
                (112, 4.49)
                (113, 4.51)
                (114, 4.56)
                (115, 4.48)
                (116, 4.44)
                (117, 4.44)
                (118, 4.46)
                (119, 4.5)
                (120, 4.5)
                (121, 4.5)
                (122, 4.46)
                (123, 4.46)
                (124, 4.48)
                (125, 4.51)
                (126, 4.45)
                (127, 4.45)
                (128, 4.43)
                (129, 4.43)
                (130, 4.42)
                (131, 4.37)
                (132, 4.36)
                (133, 4.33)
                (134, 4.32)
                (135, 4.32)
                (136, 4.31)
                (137, 4.31)
                (138, 4.29)
                (139, 4.29)
                (140, 4.33)
                (141, 4.39)
                (142, 4.33)
                (143, 4.33)
                (144, 4.27)
                (145, 4.27)
                (146, 4.28)
                (147, 4.28)
                (148, 4.24)
                (149, 4.24)
                (150, 4.25)
                (151, 4.25)
                (152, 4.36)
                (153, 4.32)
                (154, 4.32)
                (155, 4.29)
                (156, 4.2)
                (157, 4.2)
                (158, 4.25)
                (159, 4.29)
                (160, 4.24)
                (161, 4.24)
                (162, 4.27)
                (163, 4.27)
                (164, 4.2)
                (165, 4.2)
                (166, 4.21)
                (167, 4.24)
                (168, 4.16)
                (169, 4.16)
                (170, 4.24)
                (171, 4.24)
                (172, 4.26)
                (173, 4.26)
                (174, 4.27)
                (175, 4.19)
                (176, 4.19)
                (177, 4.24)
                (178, 4.26)
                (179, 4.14)
                (180, 4.14)
                (181, 4.18)
                (182, 4.13)
                (183, 4.13)
                (184, 4.18)
                (185, 4.21)
                (186, 4.19)
                (187, 4.15)
                (188, 4.15)
                (189, 4.2)
                (190, 4.2)
                (191, 4.2)
                (192, 4.2)
                (193, 4.22)
                (194, 4.23)
                (195, 4.17)
                (196, 4.17)
                (197, 4.14)
                (198, 4.14)
                (199, 4.19)
                (200, 4.21)
        };
        \legend{{\small LV-54K}}
        
        \nextgroupplot[ 
         	xtick={0, 100, 200, 300, 400},
         	ytick={10, 20, 30, 40, 50},
            xlabel={Epoch},
            legend style={at={(0.95,0.95)},legend columns=1,font=\small}
            ]
            \addplot[blue, smooth] coordinates {
                (69, 50.87)
                (70, 50.63)
                (71, 50.0)
                (72, 49.92)
                (73, 49.92)
                (74, 50.64)
                (75, 48.77)
                (76, 48.77)
                (77, 49.64)
                (78, 48.97)
                (79, 47.7)
                (80, 47.7)
                (81, 48.9)
                (82, 48.36)
                (83, 46.82)
                (84, 46.82)
                (85, 46.12)
                (86, 46.12)
                (87, 45.55)
                (88, 45.55)
                (89, 46.38)
                (90, 46.38)
                (91, 46.81)
                (92, 46.42)
                (93, 44.65)
                (94, 44.2)
                (95, 44.2)
                (96, 44.54)
                (97, 44.65)
                (98, 44.65)
                (99, 40.54)
                (100, 38.16)
                (101, 38.16)
                (102, 37.3)
                (103, 36.96)
                (104, 36.56)
                (105, 36.45)
                (106, 34.45)
                (107, 33.45)
                (108, 32.97)
                (109, 31.14)
                (110, 30.05)
                (111, 28.55)
                (112, 28.55)
                (113, 27.93)
                (114, 27.73)
                (115, 27.17)
                (116, 26.94)
                (117, 26.35)
                (118, 26.35)
                (119, 24.9)
                (120, 24.44)
                (121, 24.13)
                (122, 23.46)
                (123, 23.12)
                (124, 23.12)
                (125, 22.63)
                (126, 22.42)
                (127, 22.21)
                (128, 22.07)
                (129, 20.79)
                (130, 20.4)
                (131, 20.06)
                (132, 19.96)
                (133, 19.96)
                (134, 19.96)
                (135, 19.69)
                (136, 19.45)
                (137, 19.45)
                (138, 19.47)
                (139, 18.79)
                (140, 18.42)
                (141, 18.3)
                (142, 18.27)
                (143, 18.27)
                (144, 17.93)
                (145, 17.93)
                (146, 17.85)
                (147, 17.83)
                (148, 17.7)
                (149, 17.34)
                (150, 17.1)
                (151, 16.93)
                (152, 16.9)
                (153, 16.64)
                (154, 16.64)
                (155, 16.54)
                (156, 16.46)
                (157, 16.46)
                (158, 16.46)
                (159, 16.19)
                (160, 16.01)
                (161, 15.83)
                (162, 15.83)
                (163, 15.62)
                (164, 15.54)
                (165, 15.54)
                (166, 15.55)
                (167, 15.45)
                (168, 15.45)
                (169, 15.26)
                (170, 14.99)
                (171, 14.99)
                (172, 15.13)
                (173, 14.89)
                (174, 14.89)
                (175, 14.95)
                (176, 14.76)
                (177, 14.76)
                (178, 14.81)
                (179, 14.7)
                (180, 14.5)
                (181, 14.5)
                (182, 14.51)
                (183, 14.31)
                (184, 14.31)
                (185, 14.46)
                (186, 14.31)
                (187, 14.31)
                (188, 14.32)
                (189, 14.22)
                (190, 14.22)
                (191, 14.17)
                (192, 14.17)
                (193, 14.3)
                (194, 13.95)
                (195, 13.88)
                (196, 13.88)
                (197, 13.73)
                (198, 13.73)
                (199, 14.04)
                (200, 14.11)
                (201, 13.83)
                (202, 13.79)
                (203, 13.79)
                (204, 13.79)
                (205, 13.71)
                (206, 13.71)
                (207, 13.87)
                (208, 13.7)
                (209, 13.54)
                (210, 13.54)
                (211, 13.54)
                (212, 13.51)
                (213, 13.4)
                (214, 13.4)
                (215, 13.49)
                (216, 13.49)
                (217, 13.39)
                (218, 13.39)
                (219, 13.24)
                (220, 13.24)
                (221, 13.28)
                (222, 13.23)
                (223, 13.23)
                (224, 13.31)
                (225, 13.25)
                (226, 13.07)
                (227, 13.07)
                (228, 13.14)
                (229, 13.18)
                (230, 13.18)
                (231, 13.24)
                (232, 13.02)
                (233, 13.0)
                (234, 13.0)
                (235, 13.07)
                (236, 13.07)
                (237, 13.14)
                (238, 13.14)
                (239, 13.03)
                (240, 13.03)
                (241, 13.13)
                (242, 13.08)
                (243, 12.96)
                (244, 12.96)
                (245, 13.01)
                (246, 13.12)
                (247, 13.15)
                (248, 12.79)
                (249, 12.79)
                (250, 12.88)
                (251, 12.98)
                (252, 12.84)
                (253, 12.84)
                (254, 12.79)
                (255, 12.79)
                (256, 12.83)
                (257, 12.83)
                (258, 12.74)
                (259, 12.71)
                (260, 12.69)
                (261, 12.69)
                (262, 12.74)
                (263, 12.74)
                (264, 12.68)
                (265, 12.68)
                (266, 12.66)
                (267, 12.64)
                (268, 12.56)
                (269, 12.53)
                (270, 12.53)
                (271, 12.7)
                (272, 12.55)
                (273, 12.55)
                (274, 12.58)
                (275, 12.57)
                (276, 12.57)
                (277, 12.49)
                (278, 12.49)
                (279, 12.58)
                (280, 12.58)
                (281, 12.6)
                (282, 12.52)
                (283, 12.52)
                (284, 12.57)
                (285, 12.47)
                (286, 12.42)
                (287, 12.42)
                (288, 12.57)
                (289, 12.6)
                (290, 12.61)
                (291, 12.61)
                (292, 12.61)
                (293, 12.51)
                (294, 12.51)
                (295, 12.29)
                (296, 12.29)
                (297, 12.51)
                (298, 12.43)
                (299, 12.43)
                (300, 12.58)
                (301, 12.58)
                (302, 12.51)
                (303, 12.37)
                (304, 12.37)
                (305, 12.43)
                (306, 12.43)
                (307, 12.37)
                (308, 12.37)
                (309, 12.44)
                (310, 12.44)
                (311, 12.5)
                (312, 12.44)
                (313, 12.44)
                (314, 12.45)
                (315, 12.45)
                (316, 12.4)
                (317, 12.4)
                (318, 12.44)
                (319, 12.39)
                (320, 12.39)
                (321, 12.42)
                (322, 12.42)
                (323, 12.47)
                (324, 12.5)
                (325, 12.47)
                (326, 12.47)
                (327, 12.33)
                (328, 12.33)
                (329, 12.36)
                (330, 12.36)
                (331, 12.4)
                (332, 12.26)
                (333, 12.26)
                (334, 12.29)
                (335, 12.25)
                (336, 12.25)
                (337, 12.45)
                (338, 12.19)
                (339, 12.19)
                (340, 12.28)
                (341, 12.21)
                (342, 12.21)
                (343, 12.2)
                (344, 12.19)
                (345, 12.19)
                (346, 12.17)
                (347, 12.07)
                (348, 12.07)
                (349, 12.22)
                (350, 12.16)
                (351, 12.16)
                (352, 12.1)
                (353, 12.1)
                (354, 12.17)
                (355, 12.17)
                (356, 12.1)
                (357, 12.1)
                (358, 12.11)
                (359, 12.25)
                (360, 12.26)
                (361, 12.27)
                (362, 12.36)
                (363, 12.12)
                (364, 12.12)
                (365, 12.14)
                (366, 12.11)
                (367, 12.08)
                (368, 12.08)
                (369, 12.1)
                (370, 12.06)
                (371, 12.06)
                (372, 12.14)
                (373, 12.14)
                (374, 12.2)
                (375, 12.11)
                (376, 12.11)
                (377, 12.19)
                (378, 12.11)
                (379, 12.11)
                (380, 12.16)
                (381, 12.15)
                (382, 12.15)
                (383, 12.11)
                (384, 12.11)
                (385, 12.15)
                (386, 12.05)
                (387, 12.05)
                (388, 12.13)
                (389, 12.13)
                (390, 12.15)
                (391, 12.22)
                (392, 12.14)
                (393, 12.03)
                (394, 12.03)
                (395, 12.03)
                (396, 11.91)
                (397, 11.91)
                (398, 12.04)
                (399, 12.11)
                (400, 11.7)
        };
        \addplot[red, smooth] coordinates {
            (49, 50.17)
            (50, 50.17)
            (51, 49.35)
            (52, 49.35)
            (53, 49.9)
            (54, 48.02)
            (55, 48.02)
            (56, 48.85)
            (57, 47.69)
            (58, 47.69)
            (59, 48.05)
            (60, 48.54)
            (61, 48.89)
            (62, 48.75)
            (63, 47.27)
            (64, 47.27)
            (65, 46.9)
            (66, 46.9)
            (67, 47.26)
            (68, 46.61)
            (69, 46.61)
            (70, 47.55)
            (71, 46.5)
            (72, 45.75)
            (73, 45.75)
            (74, 45.53)
            (75, 45.53)
            (76, 46.18)
            (77, 46.18)
            (78, 46.1)
            (79, 46.1)
            (80, 45.41)
            (81, 45.41)
            (82, 46.16)
            (83, 46.25)
            (84, 46.11)
            (85, 45.21)
            (86, 45.21)
            (87, 45.42)
            (88, 45.51)
            (89, 45.88)
            (90, 44.73)
            (91, 44.35)
            (92, 44.35)
            (93, 44.56)
            (94, 44.48)
            (95, 44.48)
            (96, 44.9)
            (97, 44.9)
            (98, 44.2)
            (99, 29.18)
            (100, 27.9)
            (101, 26.55)
            (102, 24.6)
            (103, 24.6)
            (104, 25.03)
            (105, 24.03)
            (106, 24.03)
            (107, 23.43)
            (108, 23.43)
            (109, 19.75)
            (110, 19.62)
            (111, 18.69)
            (112, 18.69)
            (113, 18.68)
            (114, 18.39)
            (115, 17.89)
            (116, 17.89)
            (117, 17.91)
            (118, 18.45)
            (119, 16.16)
            (120, 16.16)
            (121, 16.12)
            (122, 16.03)
            (123, 16.03)
            (124, 16.02)
            (125, 16.01)
            (126, 15.4)
            (127, 15.4)
            (128, 15.6)
            (129, 13.43)
            (130, 12.96)
            (131, 12.69)
            (132, 12.69)
            (133, 12.76)
            (134, 12.62)
            (135, 12.62)
            (136, 12.79)
            (137, 12.78)
            (138, 12.78)
            (139, 12.92)
            (140, 12.37)
            (141, 12.37)
            (142, 12.5)
            (143, 12.5)
            (144, 12.6)
            (145, 12.6)
            (146, 12.29)
            (147, 12.17)
            (148, 12.17)
            (149, 11.37)
            (150, 11.2)
            (151, 10.82)
            (152, 10.82)
            (153, 10.87)
            (154, 10.86)
            (155, 10.86)
            (156, 10.93)
            (157, 10.87)
            (158, 10.79)
            (159, 10.54)
            (160, 10.54)
            (161, 10.62)
            (162, 10.68)
            (163, 10.68)
            (164, 10.78)
            (165, 10.78)
            (166, 10.82)
            (167, 11.0)
            (168, 10.63)
            (169, 9.96)
            (170, 9.96)
            (171, 9.6)
            (172, 9.6)
            (173, 9.67)
            (174, 9.82)
            (175, 9.65)
            (176, 9.65)
            (177, 10.08)
            (178, 10.13)
            (179, 9.79)
            (180, 9.69)
            (181, 9.69)
            (182, 9.79)
            (183, 9.69)
            (184, 9.1)
            (185, 9.05)
            (186, 9.05)
            (187, 9.05)
            (188, 8.9)
            (189, 8.9)
            (190, 9.1)
            (191, 9.1)
            (192, 9.16)
            (193, 9.08)
            (194, 8.92)
            (195, 8.92)
            (196, 9.01)
            (197, 9.01)
            (198, 9.03)
            (199, 8.57)
            (200, 8.57)
            (201, 8.68)
            (202, 8.78)
            (203, 8.78)
            (204, 8.51)
            (205, 8.48)
            (206, 8.43)
            (207, 8.4)
            (208, 8.4)
            (209, 8.56)
            (210, 8.49)
            (211, 8.49)
            (212, 8.52)
            (213, 8.72)
            (214, 8.72)
            (215, 8.63)
            (216, 8.63)
            (217, 8.64)
            (218, 8.55)
            (219, 8.55)
            (220, 8.66)
            (221, 8.66)
            (222, 8.71)
            (223, 8.77)
            (224, 8.77)
            (225, 8.58)
            (226, 8.58)
            (227, 8.64)
            (228, 8.64)
            (229, 8.31)
            (230, 8.31)
            (231, 8.34)
            (232, 8.27)
            (233, 8.27)
            (234, 8.35)
            (235, 8.46)
            (236, 8.5)
            (237, 8.5)
            (238, 8.55)
            (239, 8.55)
            (240, 8.74)
            (241, 8.49)
            (242, 8.49)
            (243, 8.5)
            (244, 8.5)
            (245, 8.53)
            (246, 8.62)
            (247, 8.47)
            (248, 8.47)
            (249, 8.26)
            (250, 8.21)
            (251, 8.03)
            (252, 8.03)
            (253, 7.96)
            (254, 7.96)
            (255, 7.95)
            (256, 7.95)
            (257, 8.0)
            (258, 8.0)
            (259, 7.95)
            (260, 7.95)
            (261, 8.04)
            (262, 8.04)
            (263, 8.07)
            (264, 7.92)
            (265, 7.92)
            (266, 7.83)
            (267, 7.83)
            (268, 7.96)
            (269, 7.94)
            (270, 7.94)
        };
        \legend{{\small LS-860}, {\small LV-54K}}
        
        \nextgroupplot[ 
         	xtick={700, 800, 900, 1000},
         	ytick={0, 20, 40, 60, 80, 100},
            xlabel={Epoch},
            legend style={at={(0.95,0.95)},legend columns=1,font=\small}
            ]
            \addplot[red, smooth] coordinates {
                (700, 77.62)
                (701, 77.36)
                (702, 77.08)
                (703, 76.8)
                (704, 76.8)
                (705, 77.1)
                (706, 77.44)
                (707, 77.38)
                (708, 77.38)
                (709, 77.6)
                (710, 77.04)
                (711, 77.04)
                (712, 76.94)
                (713, 76.94)
                (714, 77.36)
                (715, 77.54)
                (716, 76.92)
                (717, 76.61)
                (718, 76.61)
                (719, 77.17)
                (720, 77.26)
                (721, 77.26)
                (722, 77.48)
                (723, 77.13)
                (724, 77.06)
                (725, 77.06)
                (726, 76.96)
                (727, 76.63)
                (728, 76.63)
                (729, 76.93)
                (730, 76.93)
                (731, 77.61)
                (732, 77.55)
                (733, 77.43)
                (734, 76.41)
                (735, 76.41)
                (736, 77.23)
                (737, 76.47)
                (738, 76.47)
                (739, 77.06)
                (740, 76.85)
                (741, 76.85)
                (742, 77.5)
                (743, 77.55)
                (744, 77.18)
                (745, 76.97)
                (746, 76.79)
                (747, 76.79)
                (748, 77.26)
                (749, 76.26)
                (750, 76.26)
                (751, 76.24)
                (752, 76.24)
                (753, 77.07)
                (754, 76.53)
                (755, 76.53)
                (756, 76.37)
                (757, 76.37)
                (758, 76.78)
                (759, 76.73)
                (760, 76.14)
                (761, 76.14)
                (762, 77.07)
                (763, 77.75)
                (764, 77.15)
                (765, 75.93)
                (766, 75.93)
                (767, 76.38)
                (768, 77.18)
                (769, 76.67)
                (770, 76.67)
                (771, 76.44)
                (772, 76.44)
                (773, 77.07)
                (774, 76.84)
                (775, 76.48)
                (776, 76.48)
                (777, 76.51)
                (778, 75.9)
                (779, 75.9)
                (780, 76.76)
                (781, 77.0)
                (782, 77.0)
                (783, 76.72)
                (784, 76.72)
                (785, 76.72)
                (786, 77.18)
                (787, 76.29)
                (788, 76.29)
                (789, 76.5)
                (790, 76.97)
                (791, 76.97)
                (792, 76.99)
                (793, 76.78)
                (794, 76.07)
                (795, 76.07)
                (796, 76.71)
                (797, 76.71)
                (798, 77.3)
                (799, 68.44)
                (800, 65.23)
                (801, 65.07)
                (802, 64.82)
                (803, 64.82)
                (804, 64.43)
                (805, 64.43)
                (806, 63.69)
                (807, 63.69)
                (808, 64.29)
                (809, 54.38)
                (810, 54.38)
                (811, 53.42)
                (812, 53.42)
                (813, 54.23)
                (814, 54.23)
                (815, 54.03)
                (816, 53.82)
                (817, 53.82)
                (818, 53.28)
                (819, 47.25)
                (820, 46.88)
                (821, 46.01)
                (822, 46.01)
                (823, 46.95)
                (824, 45.34)
                (825, 45.34)
                (826, 46.12)
                (827, 46.12)
                (828, 46.26)
                (829, 42.36)
                (830, 41.24)
                (831, 40.92)
                (832, 40.87)
                (833, 40.44)
                (834, 40.44)
                (835, 39.62)
                (836, 39.62)
                (837, 40.01)
                (838, 40.01)
                (839, 35.93)
                (840, 35.77)
                (841, 35.77)
                (842, 35.17)
                (843, 35.17)
                (844, 36.08)
                (845, 36.08)
                (846, 35.74)
                (847, 35.74)
                (848, 35.08)
                (849, 32.35)
                (850, 32.35)
                (851, 32.29)
                (852, 32.22)
                (853, 31.84)
                (854, 31.7)
                (855, 31.7)
                (856, 31.73)
                (857, 31.74)
                (858, 31.56)
                (859, 29.39)
                (860, 29.39)
                (861, 29.66)
                (862, 29.31)
                (863, 29.31)
                (864, 29.59)
                (865, 29.3)
                (866, 29.3)
                (867, 28.94)
                (868, 28.94)
                (869, 27.22)
                (870, 27.22)
                (871, 27.38)
                (872, 26.91)
                (873, 26.91)
                (874, 27.13)
                (875, 26.3)
                (876, 26.3)
                (877, 26.33)
                (878, 26.46)
                (879, 25.15)
                (880, 25.15)
                (881, 25.45)
                (882, 24.88)
                (883, 24.88)
                (884, 25.0)
                (885, 24.89)
                (886, 24.89)
                (887, 24.92)
                (888, 24.92)
                (889, 24.67)
                (890, 24.13)
                (891, 24.01)
                (892, 24.01)
                (893, 23.8)
                (894, 23.67)
                (895, 23.65)
                (896, 23.63)
                (897, 23.63)
                (898, 23.9)
                (899, 23.35)
                (900, 23.3)
                (901, 23.23)
                (902, 23.23)
                (903, 23.22)
                (904, 23.22)
                (905, 23.07)
                (906, 22.9)
                (907, 22.9)
                (908, 22.9)
                (909, 22.9)
                (910, 22.83)
                (911, 22.75)
                (912, 22.39)
                (913, 22.39)
                (914, 22.14)
                (915, 22.1)
                (916, 22.1)
                (917, 22.35)
                (918, 22.05)
                (919, 21.87)
                (920, 21.82)
                (921, 21.6)
                (922, 21.6)
                (923, 21.82)
                (924, 21.89)
                (925, 21.86)
                (926, 21.67)
                (927, 21.67)
                (928, 21.93)
                (929, 21.54)
                (930, 21.44)
                (931, 21.44)
                (932, 21.53)
                (933, 21.42)
                (934, 21.42)
                (935, 21.42)
                (936, 21.42)
                (937, 21.49)
                (938, 21.38)
                (939, 21.38)
                (940, 21.23)
                (941, 21.23)
                (942, 21.24)
                (943, 21.46)
                (944, 21.36)
                (945, 21.25)
                (946, 21.25)
                (947, 21.25)
                (948, 21.36)
                (949, 21.12)
                (950, 20.9)
                (951, 20.88)
                (952, 20.88)
                (953, 20.94)
                (954, 20.7)
                (955, 20.7)
                (956, 21.01)
                (957, 20.89)
                (958, 20.66)
                (959, 20.66)
                (960, 20.75)
                (961, 20.82)
                (962, 20.58)
                (963, 20.58)
                (964, 20.86)
                (965, 20.87)
                (966, 20.92)
                (967, 20.62)
                (968, 20.62)
                (969, 20.66)
                (970, 20.58)
                (971, 20.51)
                (972, 20.51)
                (973, 20.18)
                (974, 20.18)
                (975, 20.36)
                (976, 20.36)
                (977, 20.36)
                (978, 20.36)
                (979, 20.18)
                (980, 20.18)
                (981, 20.34)
                (982, 20.58)
                (983, 20.39)
                (984, 20.39)
                (985, 20.43)
                (986, 20.44)
                (987, 20.44)
                (988, 20.46)
                (989, 20.49)
            };
        \addplot[blue, smooth] coordinates {
            (700, 76.93)
            (701, 76.93)
            (702, 76.94)
            (703, 76.94)
            (704, 77.66)
            (705, 76.84)
            (706, 76.84)
            (707, 76.83)
            (708, 76.83)
            (709, 77.26)
            (710, 77.26)
            (711, 77.71)
            (712, 76.26)
            (713, 76.26)
            (714, 77.44)
            (715, 76.44)
            (716, 76.44)
            (717, 78.08)
            (718, 77.95)
            (719, 77.54)
            (720, 77.19)
            (721, 77.19)
            (722, 77.59)
            (723, 77.77)
            (724, 77.37)
            (725, 77.37)
            (726, 77.93)
            (727, 78.56)
            (728, 77.07)
            (729, 77.07)
            (730, 77.04)
            (731, 77.04)
            (732, 77.36)
            (733, 77.37)
            (734, 75.88)
            (735, 75.88)
            (736, 76.17)
            (737, 77.28)
            (738, 77.28)
            (739, 77.62)
            (740, 77.62)
            (741, 77.28)
            (742, 77.28)
            (743, 77.23)
            (744, 77.23)
            (745, 76.75)
            (746, 76.75)
            (747, 76.94)
            (748, 77.05)
            (749, 77.05)
            (750, 76.61)
            (751, 76.61)
            (752, 76.69)
            (753, 76.99)
            (754, 77.46)
            (755, 77.46)
            (756, 77.0)
            (757, 76.66)
            (758, 76.66)
            (759, 76.95)
            (760, 76.95)
            (761, 77.3)
            (762, 77.12)
            (763, 76.49)
            (764, 76.49)
            (765, 76.73)
            (766, 76.69)
            (767, 76.69)
            (768, 76.59)
            (769, 76.59)
            (770, 77.07)
            (771, 77.07)
            (772, 77.45)
            (773, 77.12)
            (774, 76.95)
            (775, 76.95)
            (776, 77.23)
            (777, 77.23)
            (778, 76.96)
            (779, 76.63)
            (780, 76.16)
            (781, 76.16)
            (782, 77.12)
            (783, 77.12)
            (784, 77.06)
            (785, 76.67)
            (786, 76.67)
            (787, 76.96)
            (788, 76.96)
            (789, 76.72)
            (790, 76.71)
            (791, 76.5)
            (792, 76.5)
            (793, 76.71)
            (794, 76.71)
            (795, 76.91)
            (796, 76.89)
            (797, 76.53)
            (798, 76.22)
            (799, 70.99)
            (800, 68.14)
            (801, 65.59)
            (802, 65.11)
            (803, 64.33)
            (804, 63.07)
            (805, 63.07)
            (806, 62.27)
            (807, 62.27)
            (808, 62.07)
            (809, 55.43)
            (810, 53.32)
            (811, 52.88)
            (812, 52.88)
            (813, 53.1)
            (814, 53.14)
            (815, 51.5)
            (816, 51.5)
            (817, 51.78)
            (818, 51.74)
            (819, 44.77)
            (820, 44.77)
            (821, 45.46)
            (822, 45.59)
            (823, 45.59)
            (824, 43.7)
            (825, 43.7)
            (826, 44.11)
            (827, 44.11)
            (828, 42.66)
            (829, 38.79)
            (830, 37.84)
            (831, 37.4)
            (832, 37.4)
            (833, 37.48)
            (834, 37.04)
            (835, 37.04)
            (836, 37.65)
            (837, 36.67)
            (838, 36.67)
            (839, 34.16)
            (840, 34.16)
            (841, 33.31)
            (842, 33.31)
            (843, 33.75)
            (844, 33.75)
            (845, 33.26)
            (846, 33.26)
            (847, 33.14)
            (848, 33.14)
            (849, 31.89)
            (850, 31.71)
            (851, 31.66)
            (852, 31.59)
            (853, 31.11)
            (854, 30.48)
            (855, 30.48)
            (856, 30.93)
            (857, 30.78)
            (858, 30.78)
            (859, 29.94)
            (860, 29.86)
            (861, 29.64)
            (862, 29.46)
            (863, 29.46)
            (864, 29.18)
            (865, 29.18)
            (866, 29.32)
            (867, 29.27)
            (868, 29.26)
            (869, 29.26)
            (870, 29.3)
            (871, 28.95)
            (872, 28.95)
            (873, 29.24)
            (874, 28.9)
            (875, 28.83)
            (876, 28.83)
            (877, 28.84)
            (878, 28.84)
            (879, 28.26)
            (880, 28.26)
            (881, 28.43)
            (882, 28.29)
            (883, 28.1)
            (884, 28.1)
            (885, 28.08)
            (886, 28.08)
            (887, 28.46)
            (888, 28.07)
            (889, 27.95)
            (890, 27.88)
            (891, 27.88)
            (892, 27.89)
            (893, 27.89)
            (894, 27.93)
            (895, 27.69)
            (896, 27.69)
            (897, 27.8)
            (898, 27.62)
            (899, 27.51)
            (900, 27.25)
            (901, 27.17)
            (902, 27.13)
            (903, 27.13)
            (904, 27.22)
            (905, 27.18)
            (906, 27.18)
            (907, 27.08)
            (908, 27.08)
            (909, 27.26)
            (910, 27.22)
            (911, 27.1)
            (912, 27.09)
            (913, 27.09)
            (914, 27.08)
            (915, 26.4)
            (916, 26.4)
            (917, 26.79)
            (918, 27.02)
            (919, 26.91)
            (920, 26.91)
            (921, 27.03)
            (922, 26.83)
            (923, 26.83)
            (924, 26.81)
            (925, 26.81)
            (926, 26.76)
            (927, 26.72)
            (928, 26.72)
            (929, 26.7)
            (930, 26.7)
            (931, 26.75)
            (932, 26.8)
            (933, 26.57)
            (934, 26.57)
            (935, 26.58)
            (936, 26.58)
            (937, 26.72)
            (938, 26.64)
            (939, 26.53)
            (940, 26.53)
            (941, 26.35)
            (942, 26.33)
            (943, 26.33)
            (944, 26.51)
            (945, 26.54)
            (946, 26.54)
            (947, 26.25)
            (948, 26.2)
            (949, 26.1)
            (950, 26.1)
            (951, 26.35)
            (952, 26.18)
            (953, 26.18)
            (954, 26.21)
            (955, 26.21)
            (956, 26.18)
            (957, 26.18)
            (958, 26.31)
            (959, 26.31)
            (960, 26.29)
            (961, 26.29)
            (962, 26.19)
            (963, 26.19)
            (964, 26.14)
            (965, 26.14)
            (966, 26.14)
            (967, 26.13)
            (968, 26.13)
            (969, 26.18)
            (970, 26.4)
            (971, 26.14)
            (972, 26.14)
            (973, 26.03)
            (974, 26.03)
            (975, 26.08)
            (976, 26.08)
            (977, 26.21)
            (978, 26.11)
            (979, 25.95)
            (980, 25.95)
            (981, 25.87)
            (982, 25.87)
            (983, 26.16)
            (984, 25.82)
            (985, 25.82)
            (986, 26.1)
            (987, 25.98)
            (988, 25.98)
            (989, 26.05)
            (990, 25.91)
            (991, 25.83)
            (992, 25.83)
            (993, 25.9)
            (994, 25.83)
            (995, 25.83)
            (996, 25.87)
            (997, 25.84)
            (998, 25.79)
            (999, 25.79)
            (1000, 25.83)
            (1001, 25.83)
            (1002, 25.91)
            (1003, 25.82)
            (1004, 25.8)
            (1005, 25.8)
            (1006, 25.86)
            (1007, 25.69)
            (1008, 25.69)
            (1009, 25.75)
            (1010, 25.83)
            (1011, 25.92)
            (1012, 25.89)
            (1013, 25.85)
            (1014, 25.83)
            (1015, 25.83)
            (1016, 25.87)
            (1017, 25.91)
            (1018, 25.92)
            (1019, 25.66)
            (1020, 25.66)
            (1021, 25.86)
            (1022, 25.7)
            (1023, 25.7)
            (1024, 25.7)
            (1025, 25.78)
            (1026, 25.78)
            (1027, 25.86)
            (1028, 25.86)
            (1029, 25.89)
            (1030, 25.81)
            (1031, 25.81)
            (1032, 25.86)
            (1033, 25.86)
            (1034, 25.71)
            (1035, 25.71)
            (1036, 25.61)
            (1037, 25.61)
            (1038, 25.8)
            (1039, 25.71)
            (1040, 25.61)
            (1041, 25.61)
        };
        \legend{{\small LV-54K}, {\small LS-960}}

    \end{groupplot}
\end{tikzpicture}

%% file: figures/a2_ipl_vs_scratch.tikz
\begin{tikzpicture}
    \begin{axis}[
        width = 10.5cm, height = 6.0cm,
        ytick={10,20,30},
        xtick={0, 200000, 400000, 600000, 800000, 1000000,1200000, 1400000},
        ylabel={Word Error Rate},
        xlabel={Training Data Consumed (Hours)},
        every x tick scale label/.style={at={(xticklabel cs:0.9)},anchor=south west},
    ]
        \addplot[red, smooth] coordinates {
            (10008.900000000018, 29.18)
            (10110.000000000018, 27.9)
            (22397.900000000016, 26.55)
            (34685.80000000002, 24.6)
            (46973.70000000002, 24.6)
            (59261.60000000002, 25.03)
            (71549.50000000001, 24.03)
            (83837.40000000001, 24.03)
            (96125.3, 23.43)
            (108413.2, 23.43)
            (120701.09999999999, 19.75)
            (132989.0, 19.62)
            (146093.35, 18.69)
            (159197.7, 18.69)
            (172302.05000000002, 18.68)
            (185406.40000000002, 18.39)
            (198510.75000000003, 17.89)
            (211615.10000000003, 17.89)
            (224719.45000000004, 17.91)
            (237823.80000000005, 18.45)
            (250928.15000000005, 16.16)
            (264032.50000000006, 16.16)
            (277248.86000000004, 16.12)
            (290465.22000000003, 16.03)
            (303681.58, 16.03)
            (316897.94, 16.02)
            (330114.3, 16.01)
            (343330.66, 15.4)
            (356547.01999999996, 15.4)
            (369763.37999999995, 15.6)
            (382979.73999999993, 13.43)
            (396196.0999999999, 12.96)
            (408992.4799999999, 12.69)
            (421788.8599999999, 12.69)
            (434585.23999999993, 12.76)
            (447381.61999999994, 12.62)
            (460177.99999999994, 12.62)
            (472974.37999999995, 12.79)
            (485770.75999999995, 12.78)
            (498567.13999999996, 12.78)
            (511363.51999999996, 12.92)
            (524159.89999999997, 12.37)
            (537232.14, 12.37)
            (540576.59, 12.5)
            (543921.0399999999, 12.5)
            (547265.4899999999, 12.6)
            (550609.9399999998, 12.6)
            (553954.3899999998, 12.29)
            (557298.8399999997, 12.17)
            (560643.2899999997, 12.17)
            (563987.7399999996, 11.37)
            (567332.1899999996, 11.2)
            (579497.5799999996, 10.82)
            (591662.9699999996, 10.82)
            (603828.3599999996, 10.87)
            (615993.7499999997, 10.86)
            (628159.1399999997, 10.86)
            (640324.5299999997, 10.93)
            (652489.9199999997, 10.87)
            (664655.3099999997, 10.79)
            (676820.6999999997, 10.54)
            (688986.0899999997, 10.54)
            (701140.1199999998, 10.62)
            (713294.1499999998, 10.68)
            (725448.1799999998, 10.68)
            (737602.2099999998, 10.78)
            (749756.2399999999, 10.78)
            (761910.2699999999, 10.82)
            (774064.2999999999, 11.0)
            (786218.33, 10.63)
            (798372.36, 9.96)
            (810526.39, 9.96)
            (823426.02, 9.6)
            (836325.65, 9.6)
            (849225.28, 9.67)
            (862124.91, 9.82)
            (875024.54, 9.65)
            (888114.8200000001, 9.65)
            (901205.1000000001, 10.08)
            (914295.3800000001, 10.13)
            (927385.6600000001, 9.79)
            (940475.9400000002, 9.69)
            (953585.6400000001, 9.69)
            (966695.3400000001, 9.79)
            (979805.04, 9.69)
            (992914.74, 9.1)
            (1006024.44, 9.05)
            (1018921.5599999999, 9.05)
            (1031818.6799999999, 9.05)
            (1044715.7999999999, 8.9)
            (1048014.9899999999, 8.9)
            (1051314.18, 9.1)
            (1064461.98, 9.1)
            (1077609.78, 9.16)
            (1090757.58, 9.08)
            (1103905.3800000001, 8.92)
            (1117053.1800000002, 8.92)
            (1130220.1800000002, 9.01)
            (1143387.1800000002, 9.01)
            (1156554.1800000002, 9.03)
            (1169721.1800000002, 8.57)
            (1182888.1800000002, 8.57)
            (1195446.7000000002, 8.68)
            (1208005.2200000002, 8.78)
            (1220563.7400000002, 8.78)
            (1233122.2600000002, 8.51)
            (1245680.7800000003, 8.48)
            (1258509.3000000003, 8.43)
            (1271337.8200000003, 8.4)
            (1284166.3400000003, 8.4)
            (1296994.8600000003, 8.56)
            (1309823.3800000004, 8.49)
            (1322904.7900000003, 8.49)
            (1335986.2000000002, 8.52)
            (1349067.61, 8.72)
            (1362149.02, 8.72)
            (1375230.43, 8.63)
            (1388485.6199999999, 8.63)
            (1401740.8099999998, 8.64)
            (1414995.9999999998, 8.55)
            (1428251.1899999997, 8.55)
            (1441506.3799999997, 8.66)
            (1454571.2599999995, 8.66)
            (1467636.1399999994, 8.71)
            (1480701.0199999993, 8.77)
            (1493765.8999999992, 8.77)
            (1506830.779999999, 8.58)
            (1519047.049999999, 8.58)
            (1531263.3199999991, 8.64)
            (1543479.5899999992, 8.64)
            (1555695.8599999992, 8.31)
            (1567912.1299999992, 8.31)
            (1571193.4999999993, 8.34)
            (1574474.8699999994, 8.27)
            (1577756.2399999995, 8.27)
            (1581037.6099999996, 8.35)
            (1584318.9799999997, 8.46)
        };
        \addplot[gray, smooth] coordinates {
            (20315.420000000002, 33.0)
            (22008.280000000002, 30.96)
            (23698.63, 30.3)
            (25394.96, 29.44)
            (27087.73, 29.17)
            (28780.94, 28.45)
            (30473.55, 28.45)
            (32168.3, 28.49)
            (33859.72, 27.7)
            (35553.12, 27.7)
            (37245.950000000004, 28.13)
            (38936.18000000001, 28.13)
            (40631.87000000001, 28.07)
            (42324.69000000001, 28.07)
            (44018.22000000001, 27.9)
            (45710.850000000006, 27.84)
            (47402.130000000005, 27.84)
            (49096.240000000005, 28.02)
            (50790.61000000001, 28.4)
            (52482.01000000001, 28.72)
            (54176.05000000001, 27.97)
            (55867.94000000001, 27.97)
            (57561.38000000001, 27.67)
            (59254.88000000001, 27.67)
            (60946.51000000001, 27.82)
            (62640.19000000001, 28.23)
            (64333.00000000001, 28.42)
            (66026.25, 28.55)
            (67722.23, 27.82)
            (69410.89, 27.82)
            (71105.18, 28.08)
            (72801.34999999999, 28.2)
            (74492.20999999999, 28.2)
            (76184.63999999998, 28.64)
            (77878.37999999999, 28.55)
            (79570.51999999999, 28.22)
            (81262.51, 28.22)
            (82956.20999999999, 28.74)
            (84649.35999999999, 28.39)
            (86342.31999999999, 28.16)
            (88037.71999999999, 28.16)
            (89729.65999999999, 28.22)
            (91421.76999999999, 28.43)
            (93114.51, 28.23)
            (94806.34, 28.23)
            (96500.73999999999, 28.69)
            (98192.90999999999, 28.73)
            (99887.83999999998, 28.68)
            (101579.16999999998, 28.68)
            (103273.26999999999, 28.71)
            (104966.12999999999, 28.71)
            (106657.30999999998, 28.47)
            (108351.69999999998, 28.47)
            (110045.04999999999, 28.78)
            (111738.04, 28.12)
            (113430.78, 28.12)
            (115123.31, 28.62)
            (116814.3, 29.21)
            (118509.49, 28.64)
            (120202.15000000001, 28.64)
            (121894.46, 28.78)
            (123588.14, 28.61)
            (125281.22, 28.61)
            (126974.3, 29.14)
            (128667.17, 28.89)
            (130359.37, 28.89)
            (132054.38999999998, 29.04)
            (133746.66999999998, 29.53)
            (135440.69999999998, 29.19)
            (137134.34, 29.19)
            (138824.63, 28.73)
            (140518.62, 28.73)
            (142210.61, 28.84)
            (143902.44999999998, 28.99)
            (145596.49999999997, 28.99)
            (147290.45999999996, 29.22)
            (148983.12999999998, 29.22)
            (150675.95999999996, 29.42)
            (152367.71999999997, 29.46)
            (154061.96999999997, 29.36)
            (155755.45999999996, 29.34)
            (157448.41999999995, 29.31)
            (159140.34999999995, 29.31)
            (160833.40999999995, 29.33)
            (162526.81999999995, 29.56)
            (164221.10999999996, 29.32)
            (165913.37999999995, 29.32)
            (167607.43999999994, 29.3)
            (169299.47999999995, 29.3)
            (170992.11999999997, 29.49)
            (172685.52999999997, 29.49)
            (174378.60999999996, 29.39)
            (176070.85999999996, 29.39)
            (177765.01999999996, 29.19)
            (179457.80999999997, 29.19)
            (181149.97999999998, 29.29)
            (182843.18, 29.29)
            (184535.84, 29.69)
            (186228.98, 29.67)
            (187921.85, 29.4)
            (189614.39, 29.4)
            (191307.04, 29.33)
            (193001.05000000002, 29.24)
            (194693.32, 29.24)
            (196387.05000000002, 29.2)
            (198079.39, 29.2)
            (199773.88, 29.43)
            (201463.75, 29.43)
            (203159.15, 29.5)
            (204851.81, 29.5)
            (206546.72, 29.7)
            (208237.9, 29.47)
            (209930.66, 29.47)
            (211623.48, 29.64)
            (213317.63, 29.64)
            (215012.99, 29.53)
            (216703.44, 29.53)
            (218395.94, 29.89)
            (220090.27, 29.89)
            (221781.75999999998, 29.61)
            (223475.24999999997, 29.61)
            (225167.34999999998, 29.6)
            (226860.45999999996, 29.6)
            (228553.56999999995, 29.62)
            (230245.89999999994, 29.62)
            (231940.06999999995, 29.75)
            (233632.46999999994, 29.75)
            (235324.82999999993, 29.7)
            (237020.40999999992, 29.7)
            (238712.15999999992, 29.76)
            (240405.6499999999, 29.75)
            (242097.5899999999, 29.57)
            (243791.0499999999, 29.57)
            (245483.2799999999, 29.68)
            (247176.81999999992, 29.68)
            (248870.18999999992, 29.68)
            (250561.6199999999, 29.76)
            (252255.37999999992, 29.58)
            (253948.29999999993, 29.58)
            (255640.96999999994, 29.65)
            (257336.62999999995, 29.74)
            (259026.50999999995, 29.61)
            (260720.32999999996, 29.61)
            (262413.94999999995, 29.47)
            (264106.81999999995, 29.47)
            (265798.81999999995, 29.53)
            (267492.37999999995, 29.63)
            (269186.63999999996, 29.57)
            (270877.86999999994, 29.57)
            (272570.2799999999, 29.65)
            (274264.6699999999, 29.68)
            (275957.31999999995, 29.73)
            (277648.8599999999, 29.73)
            (279342.54999999993, 29.79)
            (281035.6499999999, 29.64)
            (282730.2699999999, 29.51)
            (284422.9399999999, 29.51)
            (286115.59999999986, 29.6)
            (287809.22999999986, 29.64)
            (289502.6299999999, 29.64)
            (291195.1499999999, 29.67)
            (292887.0099999999, 29.67)
            (294580.5799999999, 29.71)
            (296273.5999999999, 29.68)
            (297965.93999999994, 29.68)
            (299658.43999999994, 29.76)
            (301353.43999999994, 29.81)
            (303045.4199999999, 29.81)
            (304738.62999999995, 29.72)
            (306433.27999999997, 29.72)
            (308124.91, 29.83)
            (309817.06999999995, 29.9)
            (311510.44999999995, 29.76)
            (313200.8599999999, 29.76)
            (314897.37999999995, 29.83)
            (316589.12999999995, 29.79)
            (318282.36999999994, 29.79)
            (319976.63999999996, 29.86)
            (321667.85, 29.85)
            (323361.38, 29.85)
            (325053.45, 29.84)
            (326747.65, 29.81)
            (328441.10000000003, 29.81)
            (330132.02, 29.77)
            (331825.83, 29.77)
            (333519.89, 29.88)
            (335211.38, 29.86)
            (336907.35, 29.86)
            (338599.97, 29.79)
            (340292.11, 29.79)
            (341985.25, 29.75)
            (343677.3, 29.75)
            (345370.24, 29.71)
            (347062.41, 29.71)
            (348756.37999999995, 29.82)
            (350449.01999999996, 29.81)
            (352141.67999999993, 29.74)
            (353834.9699999999, 29.74)
            (355529.8499999999, 29.78)
            (357222.49999999994, 29.74)
            (358914.23999999993, 29.69)
            (360608.50999999995, 29.69)
            (362298.30999999994, 29.67)
            (363993.9099999999, 29.67)
            (365686.2299999999, 29.71)
            (367378.98999999993, 29.71)
            (369071.8499999999, 29.74)
            (370765.8899999999, 29.79)
            (372457.7299999999, 29.79)
            (374151.43999999994, 29.82)
            (375846.1099999999, 29.81)
            (377537.4099999999, 29.81)
            (379228.2199999999, 29.82)
            (380922.7099999999, 29.82)
            (382615.8699999999, 29.82)
            (384307.59999999986, 29.82)
            (386003.4999999999, 29.85)
            (387695.0699999999, 29.83)
            (389387.0599999999, 29.83)
            (391080.6899999999, 29.89)
            (392773.4599999999, 29.82)
            (394467.1599999999, 29.82)
            (396160.17999999993, 29.84)
            (397851.9799999999, 29.81)
            (399547.11999999994, 29.81)
            (401239.12999999995, 29.86)
            (402932.18999999994, 29.88)
            (404623.23999999993, 29.88)
            (406318.9599999999, 29.89)
            (408011.4599999999, 29.82)
            (409704.6899999999, 29.82)
            (411396.97999999986, 29.81)
            (413089.95999999985, 29.81)
            (414781.63999999984, 29.89)
            (416477.88999999984, 29.88)
            (418168.41999999987, 29.85)
            (419861.72999999986, 29.85)
            (421554.38999999984, 29.88)
            (423247.4299999998, 29.81)
            (424940.7799999998, 29.81)
            (426633.7399999998, 29.81)
            (428325.7899999998, 29.81)
            (430020.2899999998, 29.81)
            (431712.4199999998, 29.81)
            (433405.62999999983, 29.82)
            (435097.31999999983, 29.83)
            (436790.69999999984, 29.82)
            (438485.5499999998, 29.79)
            (440176.6699999998, 29.79)
            (441870.4799999998, 29.79)
            (443563.1499999998, 29.71)
            (445256.6499999998, 29.71)
            (446948.8399999998, 29.74)
            (448643.7699999998, 29.74)
            (450335.7599999998, 29.78)
            (452027.66999999975, 29.78)
            (453724.45999999973, 29.78)
            (455415.1199999997, 29.78)
            (457108.4399999997, 29.8)
            (458800.7699999997, 29.79)
            (460493.14999999973, 29.79)
            (462187.1899999997, 29.84)
            (463878.5399999997, 29.83)
            (465572.5699999997, 29.83)
            (467265.59999999974, 29.8)
            (468960.43999999977, 29.78)
            (470650.8899999998, 29.78)
            (472344.29999999976, 29.84)
            (474038.92999999976, 29.82)
            (475729.80999999976, 29.82)
            (477423.8999999998, 29.83)
            (479116.3999999998, 29.83)
            (480809.4599999998, 29.83)
            (482502.6099999998, 29.83)
            (484196.8599999998, 29.83)
            (485889.2099999998, 29.83)
            (487580.5699999998, 29.82)
            (489274.8499999998, 29.81)
            (490967.9699999998, 29.8)
            (492661.2399999998, 29.79)
            (494354.4399999998, 29.79)
            (496046.4799999998, 29.79)
            (497739.9999999998, 29.8)
            (499431.56999999983, 29.8)
            (501125.50999999983, 29.79)
            (502817.8099999998, 29.79)
            (504510.5999999998, 29.81)
            (506204.0599999998, 29.81)
            (507896.76999999984, 29.8)
            (509591.32999999984, 29.8)
        };
        \addplot[purple, smooth] coordinates {
            (4066.09, 29.21)
            (8123.530000000001, 24.74)
            (12172.66, 24.15)
            (16216.08, 21.81)
            (20265.76, 20.94)
            (24305.019999999997, 20.02)
            (28347.229999999996, 19.58)
            (32419.769999999997, 19.43)
            (36470.149999999994, 18.91)
            (40520.84999999999, 18.64)
            (44566.78999999999, 18.45)
            (48618.229999999996, 17.73)
            (52661.34, 17.73)
            (56697.729999999996, 18.06)
            (60756.84, 18.06)
            (64831.869999999995, 17.57)
            (68889.87999999999, 17.57)
            (72960.65, 17.62)
            (77020.14, 17.29)
            (81056.99, 17.28)
            (85084.90000000001, 17.28)
            (89133.39000000001, 16.96)
            (93141.94000000002, 16.96)
            (97223.46000000002, 17.0)
            (101270.01000000002, 16.74)
            (105312.08000000003, 16.74)
            (109348.15000000004, 17.01)
            (113393.46000000004, 17.01)
            (117464.13000000003, 17.01)
            (121492.53000000003, 16.95)
            (125557.52000000003, 16.95)
            (129587.90000000004, 16.63)
            (133625.42000000004, 16.58)
            (137686.56000000006, 16.58)
            (141727.05000000005, 16.78)
            (145806.72000000006, 16.91)
            (149844.87000000005, 16.59)
            (153908.03000000006, 16.59)
            (157960.82000000007, 16.61)
            (162011.70000000007, 16.61)
            (166050.07000000007, 16.71)
            (170089.06000000006, 16.68)
            (174152.91000000006, 16.64)
            (178207.55000000008, 16.64)
            (182241.59000000008, 16.79)
            (186311.4800000001, 16.67)
            (190391.2900000001, 16.67)
            (194450.09000000008, 16.55)
            (198505.40000000008, 16.55)
            (202555.5400000001, 16.7)
            (206600.12000000008, 16.74)
            (210630.70000000007, 16.76)
            (214663.59000000008, 16.58)
            (218701.7200000001, 16.52)
            (222747.3800000001, 16.52)
            (226786.3800000001, 16.76)
            (230818.7300000001, 16.62)
            (234845.9200000001, 16.62)
            (238905.3000000001, 16.67)
            (242969.8500000001, 16.67)
            (247048.5800000001, 16.69)
            (251107.4900000001, 16.69)
            (255142.6400000001, 16.7)
            (259180.9600000001, 16.7)
            (263215.3400000001, 16.86)
            (267272.7200000001, 16.42)
            (271315.4800000001, 16.42)
            (275368.3300000001, 16.69)
            (279443.74000000005, 16.33)
            (283483.95000000007, 16.33)
            (287546.11000000004, 16.66)
            (291615.95000000007, 16.66)
            (295662.2700000001, 16.91)
            (299699.1000000001, 16.38)
            (303743.70000000007, 16.38)
            (307798.4700000001, 16.71)
            (311822.9800000001, 16.59)
            (315888.6400000001, 16.59)
            (319922.13000000006, 16.49)
            (323996.26000000007, 16.49)
            (328040.8300000001, 16.54)
            (332066.38000000006, 16.69)
            (336124.9100000001, 16.82)
            (340181.3900000001, 16.86)
            (344239.19000000006, 16.86)
            (348302.06000000006, 16.77)
            (352343.0900000001, 16.51)
            (356394.2700000001, 16.51)
            (360477.0300000001, 16.6)
            (364548.4600000001, 16.6)
            (368592.63000000006, 16.67)
            (372624.12000000005, 16.71)
            (376681.79000000004, 16.68)
            (380709.7, 16.68)
            (384757.29000000004, 16.49)
            (388812.31000000006, 16.49)
            (392855.48000000004, 16.8)
            (396917.24000000005, 16.8)
            (400984.92000000004, 16.76)
            (405047.73000000004, 16.67)
            (409092.32000000007, 16.67)
            (413121.17000000004, 16.83)
            (417157.22000000003, 16.65)
            (421213.54000000004, 16.65)
            (425259.98000000004, 16.93)
            (429297.15, 17.06)
            (433372.4, 16.63)
            (437412.94, 16.63)
            (441461.9, 16.65)
            (445515.32, 16.65)
            (449551.67, 16.68)
            (453622.99, 16.68)
            (457683.32, 16.79)
            (461738.77, 16.55)
            (465762.45, 16.55)
            (469813.31, 16.68)
            (473873.59, 16.81)
            (477915.81, 16.81)
            (481985.0, 16.61)
            (486026.38, 16.61)
            (490087.06, 16.75)
            (494108.06, 16.83)
            (498187.11, 16.73)
            (502234.95, 16.73)
            (506248.23000000004, 17.04)
            (510254.64, 16.82)
            (514305.65, 16.67)
            (518325.25, 16.67)
        };
        \addplot[green, smooth] coordinates {
            (4066.53, 27.58)
            (8123.25, 22.81)
            (12175.03, 22.73)
            (16221.710000000001, 20.23)
            (20271.920000000002, 18.91)
            (24311.640000000003, 17.71)
            (28356.65, 17.23)
            (32430.06, 16.65)
            (36483.200000000004, 16.41)
            (40531.060000000005, 16.29)
            (44574.600000000006, 15.54)
            (48629.32000000001, 15.54)
            (52674.93000000001, 15.31)
            (56710.57000000001, 15.02)
            (60767.93000000001, 15.02)
            (64845.62000000001, 15.04)
            (68904.58000000002, 15.01)
            (72975.68000000002, 14.27)
            (77036.53000000003, 14.19)
            (81074.29000000002, 14.07)
            (85102.34000000003, 14.07)
            (89153.75000000003, 13.99)
            (93163.07000000004, 13.95)
            (97246.63000000003, 13.95)
            (101293.27000000003, 13.8)
            (105332.96000000004, 13.8)
            (109369.86000000003, 13.83)
            (113418.66000000003, 13.83)
            (117488.85000000003, 13.73)
            (121518.10000000003, 13.73)
            (125586.98000000004, 13.82)
            (129613.22000000004, 13.72)
            (133649.45000000004, 13.46)
            (137712.06000000003, 13.46)
            (141756.10000000003, 13.58)
            (145838.99000000005, 13.56)
            (149876.99000000005, 13.3)
            (153941.58000000005, 13.3)
            (157999.85000000003, 13.37)
            (162053.23000000004, 13.2)
            (166087.63000000003, 13.15)
            (170127.55000000005, 13.15)
            (174189.06000000006, 13.19)
            (178246.48000000007, 13.19)
            (182283.34000000005, 13.28)
            (186352.23000000007, 13.21)
            (190436.10000000006, 13.21)
            (194492.35000000006, 13.28)
            (198553.61000000007, 13.11)
            (202602.89000000007, 13.11)
            (206649.00000000006, 13.39)
            (210675.24000000005, 13.17)
            (214711.49000000005, 13.17)
            (218753.58000000005, 13.4)
            (222800.51000000004, 13.4)
            (226837.19000000003, 13.36)
            (230870.23000000004, 13.16)
            (234902.34000000003, 13.16)
            (238959.00000000003, 13.31)
            (243028.56000000003, 13.39)
            (247106.17, 13.36)
            (251170.92, 13.15)
            (255204.06000000003, 13.15)
            (259241.75000000003, 13.28)
            (263275.79000000004, 13.25)
            (267331.53, 13.25)
            (271379.63, 13.3)
            (275429.48, 13.03)
            (279507.97, 13.03)
            (283547.31, 13.2)
            (287608.49, 13.36)
            (291681.89, 13.33)
            (295727.71, 13.29)
            (299766.69, 13.29)
            (303818.52, 12.99)
            (307875.26, 12.9)
            (311899.99, 12.9)
            (315957.12, 13.11)
            (319994.57, 13.26)
            (324069.64, 13.26)
            (328114.59, 13.08)
            (332138.61000000004, 13.08)
            (336203.83, 13.08)
            (340260.94, 13.11)
            (344318.7, 13.11)
            (348382.5, 13.32)
            (352425.31, 13.32)
            (356478.37, 13.32)
            (360561.27999999997, 13.15)
            (364632.93, 13.15)
            (368683.85, 13.48)
            (372713.32999999996, 13.48)
            (376767.29999999993, 13.24)
            (380794.7699999999, 13.24)
            (384844.2899999999, 13.25)
            (388896.7299999999, 13.12)
            (392941.04999999993, 13.12)
            (396969.7099999999, 13.13)
            (401027.54999999993, 13.13)
            (405091.98999999993, 13.18)
            (409154.80999999994, 13.18)
            (413198.9199999999, 13.2)
            (417224.9199999999, 13.1)
            (421272.7099999999, 13.1)
            (425323.1899999999, 13.21)
            (429369.8199999999, 13.23)
            (433416.4099999999, 13.23)
            (437495.42999999993, 13.1)
            (441539.2199999999, 13.1)
            (445581.4699999999, 13.28)
            (449628.3399999999, 13.22)
            (453670.7799999999, 13.22)
            (457742.3799999999, 13.29)
            (461804.2999999999, 13.27)
            (465859.51999999984, 13.21)
            (469894.87999999983, 13.21)
            (473946.56999999983, 13.35)
            (478001.09999999986, 13.27)
            (482051.2999999999, 13.22)
            (486123.7499999999, 12.96)
            (490165.0099999999, 12.96)
            (494231.5699999999, 12.99)
            (498250.3999999999, 13.25)
            (502325.3099999999, 13.66)
            (506366.5199999999, 13.19)
            (510381.3799999999, 13.08)
            (514385.5499999999, 13.08)
            (518438.95999999985, 13.04)
            (522468.8699999998, 13.04)
            (526521.0799999998, 13.22)
            (530581.3599999999, 13.4)
            (534629.8299999998, 13.26)
            (538680.5099999999, 13.26)
            (542772.9899999999, 13.08)
            (546823.7499999999, 13.08)
            (550883.8699999999, 13.09)
            (554938.8599999999, 13.29)
            (559005.2899999999, 13.31)
            (563055.21, 13.31)
            (567093.9199999999, 13.07)
            (571149.1599999999, 13.07)
            (575210.6799999999, 13.24)
            (579256.1199999999, 13.24)
            (583295.6399999999, 13.26)
            (587346.8599999999, 13.26)
            (591383.8799999999, 13.12)
            (595436.8599999999, 13.12)
            (599500.8499999999, 13.19)
            (603549.9599999998, 13.07)
            (607594.6499999998, 13.07)
            (611675.8999999998, 13.3)
            (615732.3099999998, 13.06)
            (619768.6399999998, 13.06)
            (623803.1099999998, 13.29)
            (627834.0699999997, 13.22)
            (631894.1399999997, 13.22)
            (635942.3099999997, 13.43)
            (640004.3899999997, 13.3)
        };
        \addplot[blue, smooth] coordinates {
            (4073.31, 28.48)
            (8135.74, 22.2)
            (12177.43, 21.82)
            (16220.37, 20.28)
            (20269.3, 18.41)
            (24311.93, 17.6)
            (28372.14, 16.44)
            (32439.12, 16.09)
            (36488.69, 15.21)
            (40537.630000000005, 15.01)
            (44587.61000000001, 14.99)
            (48642.23000000001, 14.23)
            (52682.37000000001, 14.06)
            (56707.38000000001, 14.06)
            (60773.140000000014, 13.76)
            (64829.210000000014, 13.51)
            (68890.73000000001, 13.32)
            (72970.42000000001, 13.08)
            (77028.78000000001, 13.01)
            (81051.35000000002, 12.83)
            (85105.56000000003, 12.77)
            (89128.83000000003, 12.77)
            (93180.93000000004, 12.76)
            (97235.95000000004, 12.63)
            (101268.89000000004, 12.43)
            (105321.20000000004, 12.35)
            (109377.93000000004, 12.31)
            (113452.81000000004, 12.18)
            (117499.93000000004, 11.99)
            (121550.77000000003, 11.99)
            (125578.72000000003, 12.04)
            (129620.38000000003, 11.99)
            (133694.39000000004, 11.99)
            (137762.88000000003, 11.93)
            (141809.57000000004, 11.92)
            (145868.92000000004, 11.92)
            (149903.53000000003, 11.97)
            (153960.10000000003, 11.97)
            (158009.71000000002, 11.7)
            (162050.31000000003, 11.68)
            (166085.46000000002, 11.68)
            (170148.55000000002, 11.66)
            (174198.67, 11.66)
            (178243.32, 11.75)
            (182293.87, 11.64)
            (186342.84, 11.64)
            (190431.65, 11.66)
            (194475.15, 11.66)
            (198534.78, 11.66)
            (202570.41, 11.57)
            (206588.34, 11.51)
            (210664.73, 11.51)
            (214714.28, 11.47)
            (218743.49, 11.45)
            (222794.63999999998, 11.45)
            (226820.49, 11.33)
            (230847.83, 11.33)
            (234915.31999999998, 11.59)
            (238979.11999999997, 11.2)
            (243043.96999999997, 11.2)
            (247098.76999999996, 11.77)
            (251140.60999999996, 11.63)
            (255182.94999999995, 11.53)
            (259222.05999999994, 11.43)
            (263299.6099999999, 11.43)
            (267346.5299999999, 11.31)
            (271401.8799999999, 11.31)
            (275464.7599999999, 11.44)
            (279516.22999999986, 11.44)
            (283590.9899999999, 11.4)
            (287657.3899999999, 11.4)
            (291696.7899999999, 11.21)
            (295746.87999999995, 11.21)
            (299783.2299999999, 11.29)
            (303823.5999999999, 11.29)
            (307849.7299999999, 11.24)
            (311897.8599999999, 11.24)
            (315934.4599999999, 11.33)
            (320037.5399999999, 11.33)
            (324089.56999999995, 11.32)
            (328113.79999999993, 11.28)
            (332175.69999999995, 11.28)
            (336223.76999999996, 11.26)
            (340304.51999999996, 11.15)
            (344351.36, 11.15)
            (348426.98, 11.17)
            (352454.44999999995, 11.32)
            (356552.6699999999, 11.19)
            (360605.4199999999, 11.19)
            (364634.3599999999, 11.24)
            (368678.0399999999, 11.3)
            (372739.6699999999, 11.24)
            (376783.50999999995, 11.18)
            (380808.45999999996, 11.18)
            (384842.05999999994, 11.31)
            (388886.12999999995, 11.13)
            (392955.36999999994, 11.13)
            (397021.2199999999, 11.29)
            (401075.2099999999, 11.08)
            (405107.5999999999, 11.08)
            (409154.5899999999, 11.16)
            (413191.54999999993, 11.15)
            (417265.8999999999, 11.15)
            (421303.0399999999, 11.24)
            (425323.6699999999, 11.24)
            (429404.7199999999, 11.23)
            (433446.54999999993, 11.23)
            (437497.05999999994, 11.23)
            (441561.89999999997, 11.17)
            (445591.01999999996, 11.17)
            (449666.79, 11.13)
            (453730.24, 11.13)
            (457794.04, 11.13)
            (461826.79, 11.13)
            (465880.52999999997, 10.95)
            (469929.26999999996, 10.95)
            (473984.17, 11.1)
            (478048.01999999996, 11.23)
            (482101.67999999993, 11.12)
            (486175.3599999999, 11.12)
            (490200.5199999999, 11.15)
            (494255.8499999999, 11.15)
            (498293.80999999994, 11.1)
            (502293.07999999996, 11.08)
            (506304.13999999996, 11.08)
            (510344.68999999994, 11.03)
            (514385.43999999994, 11.03)
            (518441.75999999995, 11.12)
            (522520.35, 11.18)
            (526567.26, 11.01)
            (530635.6900000001, 11.01)
            (534728.6000000001, 11.16)
            (538776.5200000001, 11.03)
            (542811.2900000002, 11.03)
            (546865.4600000002, 11.22)
            (550934.4200000002, 11.17)
            (554966.4700000002, 11.17)
            (559027.5600000002, 11.11)
            (563072.7100000002, 11.09)
            (567143.6000000002, 11.09)
            (571168.8800000002, 11.13)
            (575222.7300000002, 11.38)
            (579268.8700000002, 11.29)
            (583326.5900000002, 11.2)
            (587397.4100000001, 11.19)
            (591427.7700000001, 11.11)
            (595494.4800000001, 11.11)
            (599548.6000000001, 11.17)
            (603605.0700000001, 11.14)
            (607659.01, 11.0)
        };
    \legend{{\small IPL}, {\small Round 0}, {\small Round 1}, {\small Round 2}, {\small Round 3}}
\end{axis}
\end{tikzpicture}